\documentclass{article}
\PassOptionsToPackage{numbers,compress}{natbib}

\usepackage[final]{nips_2018}

\usepackage[utf8]{inputenc} % allow utf-8 input
\usepackage[T1]{fontenc}    % use 8-bit T1 fonts
\usepackage{hyperref}       % hyperlinks
\usepackage{url}            % simple URL typesetting
\usepackage{booktabs}       % professional-quality tables
\usepackage{amsfonts}       % blackboard math symbols
\usepackage{nicefrac}       % compact symbols for 1/2, etc.
\usepackage{microtype}      % microtypography
\usepackage{graphicx}
\usepackage{subfigure}
\usepackage{amsmath}
\usepackage{amssymb}
\usepackage{algorithm}
\usepackage{algorithmic}
\usepackage{bbm}
\usepackage{units}
\usepackage{enumitem} % for enumerations with less space
\usepackage{bm}
\usepackage{balance}
\usepackage{xcolor}
\usepackage{multirow}
\usepackage{appendix}

\newcommand\name{BRUNO}
\newcommand\fullname{Bayesian RecUrrent Neural mOdel}
\newcommand\ie{i.\,e.\ }

\DeclareSymbolFont{extraup}{U}{zavm}{m}{n}
\DeclareMathSymbol{\club}{\mathalpha}{extraup}{84}
\DeclareMathSymbol{\spade}{\mathalpha}{extraup}{85}
\DeclareMathSymbol{\heart}{\mathalpha}{extraup}{86}
\DeclareMathSymbol{\diamond}{\mathalpha}{extraup}{87}

\newlength{\mylen}
\setbox1=\hbox{$\bullet$}\setbox2=\hbox{\tiny$\bullet$}
\newcommand\rurl[1]{%
  \href{http://#1}{\nolinkurl{#1}}%
}

\def\colspaceS{2.25mm}

\def\colspaceL{4.25mm}

\def\t#1{#1}
\def\b#1{\t{\textbf{#1}}}

\newcommand\abbr[1]{\textsc{#1}}

\title{BRUNO: A Deep Recurrent Model for Exchangeable Data}

\author{
Iryna Korshunova $^\heart$\\
Ghent University \\
\texttt{iryna.korshunova@ugent.be}
\And
Jonas Degrave $^\heart\;^\dagger$\\
Ghent University \\
\texttt{jonas.degrave@ugent.be}
\And
Ferenc Husz\'{a}r\\
Twitter \\
\texttt{fhuszar@twitter.com }
\AND
Yarin Gal \\
University of Oxford \\
\texttt{yarin@cs.ox.ac.uk }
\And
Arthur Gretton $^\spadesuit$\\
Gatsby Unit, UCL \\
\texttt{arthur.gretton@gmail.com }
\And
Joni Dambre $^\spadesuit$\\
Ghent University \\
\texttt{joni.dambre@ugent.be}
}

\begin{document}
    \maketitle

    \begin{abstract}
        We present a novel model architecture which leverages deep learning tools to perform exact Bayesian inference on sets of high dimensional, complex observations. Our model is provably exchangeable, meaning that the joint distribution over observations is invariant under permutation: this property lies at the heart of Bayesian inference. The model does not require variational approximations to train, and new samples can be generated conditional on previous samples, with cost linear in the size of the conditioning set. The advantages of our architecture are demonstrated on learning tasks that require generalisation from short observed sequences while modelling sequence variability, such as conditional image generation, few-shot learning, and anomaly detection.
    \end{abstract}

    \section{Introduction}
    \label{intro}

    We address the problem of modelling unordered sets of objects that have some characteristic in common. Set modelling has been a recent focus in machine learning, both due to relevant application domains and to efficiency gains when dealing with groups of objects~\cite{edwards17, SzaSriPocGre16,vinyals16b, zaheer17}. The relevant concept in statistics is the notion of an exchangeable sequence of random variables -- a sequence where any re-ordering of the elements is equally likely. To fulfil this definition, subsequent observations must behave like previous ones, which implies that we can make predictions about the future. This property allows the formulation of some machine learning problems in terms of modelling exchangeable data. For instance, one can think of few-shot concept learning as learning to complete short exchangeable sequences~\cite{lake15}. A related example comes from the generative image modelling field, where we might want to generate images that are in some ways similar to the ones from a given set. At present, however, there are few flexible and provably exchangeable deep generative models to solve this problem.

    Formally, a finite or infinite sequence of random variables $x_1, x_2, x_3, \dots$
    is said to be exchangeable if for all $n$ and all permutations $\pi$
    \begin{align}
        p(x_1, \dots ,x_n)=p\left(x_{\pi(1)},\dots,x_{\pi(n)}\right),
    \end{align}
    \ie the joint probability remains the same under any permutation of the sequence.
    If random variables in the sequence are independent and identically distributed (i.\,i.\,d.),
    then it is easy to see that the sequence is exchangeable. The converse is false:
    exchangeable random variables can be correlated. One example of an exchangeable but non-i.\,i.\,d.\ sequence is a sequence of variables
    $x_1, \dots, x_n$, which jointly have a multivariate normal distribution $\mathcal{N}_n(\bm 0, \bm \Sigma)$
    with the same variance and covariance for all the dimensions~\cite{aldous}: $\Sigma_{ii} = 1  \textnormal{ and } \Sigma_{ij, i\ne j}= \rho, \textnormal{ with } 0 \le \rho < 1$.

    The concept of exchangeability is intimately related to Bayesian statistics.
    De Finetti's theorem states that every exchangeable process (infinite sequence of random variables) is a mixture of i.\,i.\,d.\ processes:
    \begin{equation}
        \label{eq:1}
        p(x_1,\dots,x_n)= \int p(\theta) \prod_{i=1}^n{p(x_i|\theta)d\theta},
    \end{equation}
    where $\theta$ is some parameter (finite or infinite dimensional) conditioned on which, the random variables are i.\ i.\ d.~\cite{aldous}. In our previous Gaussian example, one can prove that $x_1, \dots, x_n$ are i.\,i.\,d.\ with $x_i \sim \mathcal{N}(\theta, 1 - \rho)$ conditioned on $\theta \sim \mathcal{N}(0, \rho)$.

    In terms of predictive distributions $p(x_{n}|x_{1:n-1})$, the stochastic process in Eq.~\ref{eq:1}  can be written as
    \begin{equation}
        \label{eq:2}
        p(x_n|x_{1:n-1})= \int p(x_{n}|\theta)p(\theta|x_{1:n-1})d\theta,
    \end{equation}
    by conditioning both sides on $x_{1:n-1}$. Eq.~\ref{eq:2} is exactly the posterior predictive distribution, where we
    marginalise the likelihood of $x_{n}$ given $\theta$  with respect to the posterior distribution of $\theta$.
    From this follows one possible interpretation of the de Finetti's theorem: learning to fit an exchangeable model to sequences of data is implicitly the same as learning to reason about the hidden variables behind the data.

    One strategy for defining models of exchangeable sequences is through explicit Bayesian modelling: one defines a prior $p(\theta)$, a likelihood $p(x_i|\theta)$ and calculates the posterior in Eq.~\ref{eq:1} directly. Here, the key difficulty is the intractability of the posterior and the predictive distribution $p(x_n|x_{1:n-1})$. Both of these expressions require integrating over the parameter $\theta$, so we might end up having to use approximations. This could violate the exchangeability property and make explicit Bayesian modelling difficult.

    On the other hand, we do not have to explicitly represent the posterior to ensure exchangeability. One could define a predictive distribution $p(x_n|x_{1:n-1})$ directly, and as long as the process is exchangeable, it is consistent with Bayesian reasoning. The key difficulty here is defining an easy-to-calculate $p(x_n|x_{1:n-1})$ which satisfies exchangeability. For example, it is not clear how to train or modify an ordinary recurrent neural network (RNN) to model exchangeable data. In our opinion, the main challenge is to ensure that a hidden state contains information about all previous inputs $x_{1:n}$ regardless of sequence length.

    In this paper, we propose a novel architecture which combines features of the approaches above, which we will refer to as \name: \fullname.
    Our model is {\em provably exchangeable}, and makes use of deep features learned from observations so as to model complex data types such as images. To achieve this, we construct a {\em bijective} mapping between random variables $x_i \in \mathcal X$ in the observation space and features $z_i \in \mathcal Z$, and explicitly define an exchangeable model for the sequences $z_1, z_2, z_3, \dots$, where we know an analytic form of $p(z_n|z_{1:n-1})$ without explicitly computing the integral in Eq.~\ref{eq:2}.

    Using \name, we are able to generate samples conditioned on the input sequence by sampling directly from $p(x_n|x_{1:n-1} )$. The latter is also tractable to evaluate, \ie has linear complexity in the number of data points. %In particular, we never need to infer the posterior over some meaningful latent variable.
    In respect of model training, evaluating the predictive distribution requires a single pass through the neural network that implements $\mathcal X \mapsto \mathcal Z$ mapping. The model can be learned straightforwardly, since $p(x_n|x_{1:n-1})$ is differentiable with respect to the model parameters.

    The paper is structured as follows. In Section~\ref{related_work} we will look at two methods selected to highlight the relation of our work with previous approaches to modelling exchangeable data. Section~\ref{sec:method} will describe \name, along with necessary background information. In Section~\ref{sec:experiments}, we will use our model for conditional image generation, few-shot learning, set expansion and set anomaly detection. Our code is available at \rurl{github.com/IraKorshunova/bruno}.

    \section{Related work}
    \label{related_work}

    Bayesian sets~\cite{ghahramani06} aim to model exchangeable sequences of binary random variables by analytically computing the integrals in Eq.~\ref{eq:1},~\ref{eq:2}. This is made possible by using a Bernoulli distribution for the likelihood and a beta distribution for the prior. To apply this method to other types of data, e.g. images, one needs to engineer a set of binary features~\cite{heller06}. In that case, there is usually no one-to-one mapping between the input space $\mathcal X$ and the features space $\mathcal Z$: in consequence, it is not possible to draw samples from $p(x_n|x_{1:n-1})$.
    Unlike Bayesian sets, our approach does have a bijective transformation, which guarantees that inference in $\mathcal Z$ is equivalent to inference in space $\mathcal X$.

    The neural statistician~\cite{edwards17} is an extension of a variational autoencoder model~\cite{kingma14b, rezende14} applied to datasets. In addition to learning an approximate inference network over the latent variable $\bm z_i$ for every $\bm x_i$ in the set, approximate inference is also implemented over a latent variable $\bm c$ -- a context that is global to the dataset. The architecture for the inference network $q(\bm c | \bm x_1,\dots,\bm x_n)$ maps every $\bm x_i$ into a feature vector and applies a mean pooling operation across these representations. The resulting vector is then used to produce parameters of a Gaussian distribution over $\bm c$. Mean pooling makes $q(\bm c | \bm x_1,\dots,\bm x_n)$ invariant under permutations of the inputs. In addition to the inference networks, the neural statistician also has a generative component $p(\bm x_1,\dots,\bm x_n|\bm c)$ which assumes that $\bm x_i$'s are independent given $\bm c$. Here, it is easy to see that $\bm c$ plays the role of $\theta$ from Eq.~\ref{eq:1}. In the neural statistician, it is intractable to compute $p(\bm x_1,\dots,\bm x_n)$, so its variational lower bound is used instead. In our model, we perform an implicit inference over $\theta$ and can exactly compute predictive distributions and the marginal likelihood. Despite these differences, both neural statistician and \name\space can be applied in similar settings, namely few-shot learning and conditional image generation, albeit with some restrictions, as we will see in Section~\ref{sec:experiments}.

    \section{Method}
    \label{sec:method}

    %% AG: note: I don't say we introduce the ``exchangeable'' student t process in the first section, since initially
    % the process is not exchangeable (which is why the cost is cubic...)

\begin{figure*}[htb]
        \centerline{\includegraphics[scale=0.8]{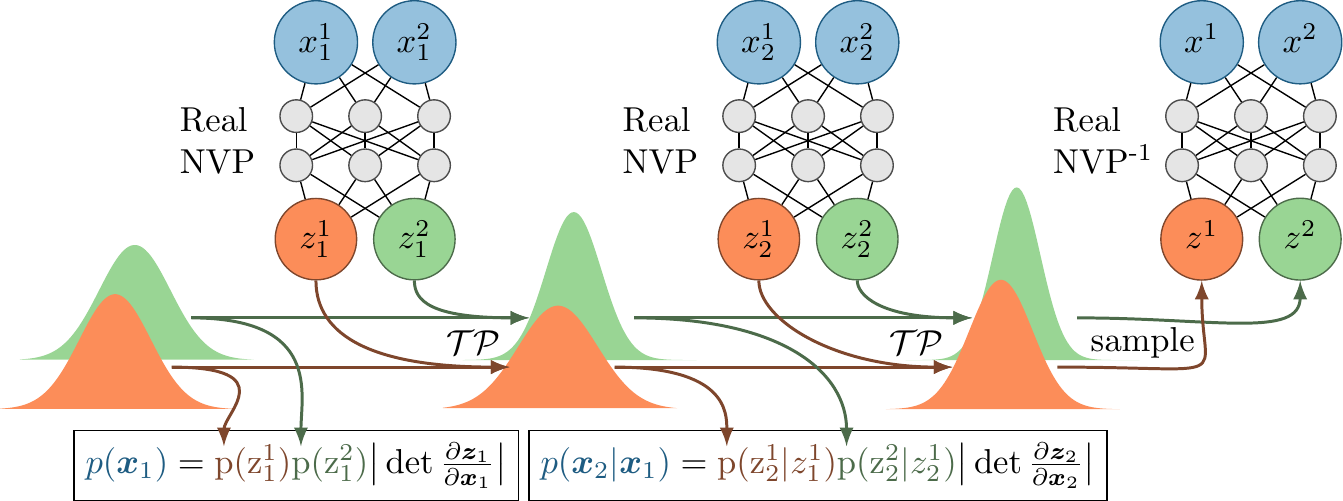}}
        \caption{A schematic of the \name\space model. It depicts how Bayesian thinking can lead to an RNN-like computational graph in which Real NVP is a bijective feature extractor and the recurrence is represented by Bayesian updates of an exchangeable Student-t process.}
        \label{fig:0}
        \vskip -0.1in
    \end{figure*}

    We begin this section with an overview of the mathematical tools needed to construct our model: first the
    Student-t process~\cite{shah14}; and then the Real NVP -- a deep, stably invertible and learnable neural network architecture for density estimation~\cite{dinh17}. We next propose \name, wherein we combine an exchangeable Student-t process with the Real NVP, and derive recurrent equations for the predictive distribution such that our model can be trained as an RNN.
    Our model is illustrated in Figure~\ref{fig:0}.

    \subsection{Student-t processes}

    The Student-t process ($\mathcal{TP}$) is the most general elliptically symmetric process with an analytically representable density~\cite{shah14}. The more commonly used Gaussian processes ($\mathcal GP$s) can be seen as limiting case of  $\mathcal{TP}$s. In what follows, we provide the background and definition of $\mathcal{TP}$s.

    Let us assume that $\bm z = (z_1, \dots z_n) \in \mathbb{R}^n$ follows a multivariate Student-t distribution $MVT_n(\nu, \bm \mu, \bm K)$ with
    degrees of freedom $\nu \in \mathbb{R}_+ \setminus[0,2] $, mean $\bm \mu \in \mathbb{R}^n$ and a positive definite $n \times n$
    covariance matrix $\bm K$. Its density is given by%
    \begin{equation}
        \label{eq:pdf}
        p(\bm z) = \frac{\Gamma(\frac{\nu+n}{2})}{((\nu-2)\pi)^{n/2}\Gamma(\nu/2)}| \bm K|^{-1/2} \bigg( 1+\frac{(\bm z- \bm \mu)^T \bm K^{-1}( \bm z- \bm \mu)}{\nu-2} \bigg) ^{-\frac{\nu+n}{2}}.
    \end{equation}
    For our problem, we are interested in computing a conditional distribution. Suppose we can partition $\bm z$ into two consecutive parts ${\bm z_a \in \mathbb{R}^{n_a}}$ and ${\bm z_b \in \mathbb{R}^{n_b}}$, such that
    \begin{equation}
        \begin{bmatrix}
            \bm z_a \cr \bm z_b
        \end{bmatrix} \sim MVT_n\Bigg(\nu, \begin{bmatrix}
                                               \bm \mu_a \cr \bm \mu_b
        \end{bmatrix},
        \begin{bmatrix}
            \bm K_{aa} & \bm K_{ab} \cr
            \bm K_{ba} & \bm K_{bb}
        \end{bmatrix}\Bigg).
    \end{equation}
    Then conditional distribution $p(\bm z_b| \bm z_a)$ is given by
    \begin{equation}
        \label{eq:3}
        \begin{split}
            &p(\bm z_b| \bm z_a) = MVT_{n_b}\Big(\nu + n_a, \bm{\tilde{\mu}_b}, \frac{\nu + \beta_a -2}{\nu + n_a-2} \bm{\tilde K}_{bb}\Big), \\
            &\bm{\tilde{\mu}_b} = \bm K_{ba}\bm K_{aa}^{-1}(\bm z_a - \bm \mu_a) + \bm \mu_b\\
            &\beta_a = (\bm z_a-\bm \mu_a)^T \bm K_{aa}^{-1} (\bm z_a - \bm \mu_a)\\
            &\bm{\tilde K}_{bb} = \bm K_{bb} - \bm K_{ba} \bm K_{aa}^{-1} \bm K_{ab}.
        \end{split}
    \end{equation}
    In the general case, when one needs to invert the covariance matrix, the complexity of computing $p(\bm z_b| \bm z_a)$ is $\mathcal O(n_a^3)$. These computations become infeasible for large datasets, which is a known bottleneck for $\mathcal GP$s and $\mathcal TP$s~\cite{rasmussen05}. In Section~\ref{sec:model}, we will show that exchangeable processes do not have this issue.

    The parameter $\nu$, representing the degrees of freedom, has a large impact on the behaviour of $\mathcal TP$s.
    It controls how heavy-tailed the t-distribution is: as $\nu$ increases, the tails get lighter and the t-distribution gets closer to the Gaussian. From Eq.~\ref{eq:3}, we can see that as $\nu$ or $n_a$ tends to infinity, the predictive distribution tends to the one from a $\mathcal GP$. Thus, for small $\nu$ and $n_a$, a $\mathcal TP$ would give less certain predictions than its corresponding $\mathcal GP$.

    A second feature of the $\mathcal TP$ is the scaling of the predictive variance with a $\beta_a$ coefficient, which explicitly depends on the values of the conditioning observations. From Eq.~\ref{eq:3}, the value of $\beta_a$ is precisely the Hotelling statistic for the vector $\bm z_a$, and has a $\chi^2_{n_a}$ distribution with mean $n_a$ in the event that $\bm z_a \sim \mathcal N_{n_a} (\bm \mu_a, \bm K_{aa})$. Looking at the weight $\nicefrac{(\nu + \beta_a - 2)}{(\nu + n_a - 2)}$, we see that the variance of $p(\bm z_b|\bm z_a)$ is increased over the Gaussian default when $\beta_a > n_a$, and is reduced otherwise. In other words, when the samples are dispersed more than they would be under the Gaussian distribution, the predictive uncertainty is increased compared with the Gaussian case.
    It is helpful in understanding these two properties to recall that the multivariate Student-t distribution can be thought of as a Gaussian distribution with an inverse Wishart prior on the covariance \cite{shah14}.
    %We return to this point in our experiments of Section~\ref{section:gp}, where this prior ensures stability when the covariance parameters of the $\mathcal TP$ model are fit from data, as compared with a $\mathcal GP$ model.

    \subsection{Real NVP}

    Real NVP~\cite{dinh17} is a member of the normalising flows family of models, where some density in the input space $\mathcal X$ is transformed into a desired probability distribution in space $\mathcal Z$ through a sequence of invertible mappings~\cite{rezende15}. Specifically, Real NVP proposes a design for a bijective function $f: \mathcal{X} \mapsto \mathcal{Z}$ with $\mathcal{X}=\mathbb{R}^D$ and $\mathcal{Z}=\mathbb{R}^D$
    such that \textbf{(a)} the inverse is easy to evaluate, i.e. the cost of computing $\bm x = f^{-1}(\bm z)$ is the same as for the forward mapping,
    and \textbf{(b)} computing the Jacobian determinant takes linear time in the number of dimensions~$D$.
    Additionally, Real NVP assumes a simple distribution for $\bm z$, e.g. an isotropic Gaussian, so one can use a change of variables formula to evaluate $p(\bm x)$:%
    \begin{equation}
        \label{eq:4}
        p(\bm x) = p(\bm z) \left| \det \Bigg( \frac{\partial f(\bm x)}{ \partial \bm x} \Bigg)\right|.
    \end{equation}
    %Moreover, sampling from $p(\bm x)$ is equivalent to sampling $\bm z$ and computing $\bm x = f^{-1}(\bm z)$.
    %
    The main building block of Real NVP is a coupling layer. It implements a mapping $\mathcal X \mapsto \mathcal Y$ that transforms half of its inputs while copying the other half directly to the output:
    \begin{equation}
        \label{eq:9}
        \begin{cases}
            \bm y^{1:d} = \bm x^{1:d} \\
            \bm y^{d+1:D} = \bm x^{d+1:D} \odot \textnormal{exp}(s(\bm x^{1:d})) + t(\bm x^{1:d}),
        \end{cases}
    \end{equation}
    where $\odot$ is an elementwise product, $s$ (scale) and $t$ (translation) are arbitrarily complex functions, e.g. convolutional neural networks.

    One can show that the coupling layer is a bijective, easily invertible mapping with a triangular Jacobian
    and composition of such layers preserves these properties.
    To obtain a highly nonlinear mapping $f(\bm x)$, one needs to stack coupling layers $\mathcal X \mapsto \mathcal Y_1 \mapsto \mathcal Y_2 \dots \mapsto \mathcal Z$ while alternating the dimensions that are being copied to the output.

    To make good use of modelling densities, the Real NVP has to treat its inputs as instances of a continuous random variable~\cite{theis16}. To do so, integer pixel values in $\bm x$ are dequantised by adding uniform noise ${\bm u \in [0,1)^D}$. The values ${\bm x + \bm u \in [0, 256)^D}$ are then rescaled to a $[0,1)$ interval and transformed with an elementwise function: ${f(x)=\textnormal{logit}(\alpha + (1 - 2 \alpha)x)}$ with some small $\alpha$.

    \subsection{\name: the exchangeable sequence model}
    \label{sec:model}

    We now combine Bayesian and deep learning tools from the previous sections and present our model for exchangeable sequences whose schematic is given in Figure~\ref{fig:0}.

    Assume we are given an exchangeable sequence $\bm x_1,\dots, \bm x_n$, where every element is a D-dimensional vector: ${\bm x_i = (x^1_i, \dots x^D_i)}$. We apply a Real NVP transformation to every $\bm x_i$, which results in an exchangeable sequence in the latent space: $\bm z_1, \dots, \bm z_n$, where ${\bm z_i \in \mathbb{R}^D}$. The proof that the latter sequence is exchangeable is given in Appendix A.

    We make the following assumptions about the latents:

    \textbf{A1}: dimensions $\{z^d\}_{d=1,\dots,D}$ are independent, so ${p(\bm z) = \prod_{d=1}^D p(z^d)}$

    \textbf{A2}: for every dimension $d$, we assume the following:
    $(z^d_1, \dots z^d_n) \sim  MVT_n (\nu^d, \mu^d \bm 1, \bm K^d)$, with parameters:

    \begin{itemize}
 \item degrees of freedom $\nu^d \in \mathbb{R}_+ \setminus[0,2] $

    \item  mean $\mu^d \bm 1$ is a $1 \times n$ dimensional vector of ones multiplied by the scalar $\mu^d \in \mathbb{R}$

    \item $n \times n$ covariance matrix $\bm K^d$ with $\bm K^d_{ii} = v^d$ and $\bm K^d_{ij, i \ne j} = \rho^d$
    where $0 \le \rho^d < v^d$ to make sure that $\bm K^d$ is a positive-definite matrix that complies
    with covariance properties of exchangeable sequences~\cite{aldous}.
\end{itemize}

%    \textbf{1)} degrees of freedom $\nu^d \in \mathbb{R}_+ \setminus[0,2] $
%
%    \textbf{2)}  mean $\mu^d \bm 1$ is a $1 \times n$ dimensional vector of ones multiplied by the scalar $\mu^d \in \mathbb{R}$
%
%    \textbf{3)} $n \times n$ covariance matrix $\bm K^d$ with $\bm K^d_{ii} = v^d$ and $\bm K^d_{ij, i \ne j} = \rho^d$
%    where $0 \le \rho^d < v^d$ to make sure that $\bm K^d$ is a positive-definite matrix that complies
%    with covariance properties of exchangeable sequences~\cite{aldous}.

    The exchangeable structure of the covariance matrix and having the same mean for every $n$, guarantees that the sequence $z^d_1, z^d_2 \dots z^d_n$ is exchangeable. Because the covariance matrix is simple, we can derive recurrent updates for the parameters of $p(z_{n+1}^d|z_{1:n}^d)$. Using the recurrence is a lot more efficient compared to the closed-form expressions in Eq.~\ref{eq:3} since we want to compute the predictive distribution for every step $n$.

    We start from a prior Student-t distribution for $p(z_1)$ with parameters $\mu_1 = \mu$ , $v_1 = v$, $\nu_{1} = \nu$, $\beta_{1} = 0$. Here, we will drop the dimension index $d$ to simplify the notation. A detailed derivation of the following results is given in Appendix B. To compute the degrees of freedom, mean and variance of $p(z_{n+1}|z_{1:n})$ for every $n$, we begin with the recurrent relations
    \begin{equation}
        \label{eq:12}
        \nu_{n+1} = \nu_n + 1, \quad \mu_{n+1} =  (1-d_n)\mu_{n} + d_n z_n, \quad v_{n+1} =  (1-d_n) v_{n} + d_n (v - \rho), \\
    \end{equation}
    where $d_{n} = \frac{\rho}{v + \rho(n-1)}$. Note that the $\mathcal GP$ recursions simply use the latter two equations, i.e. if we were to assume that ${(z^d_1, \dots z^d_n) \sim  \mathcal N_n (\mu^d \bm 1, \bm K^d)}$. For $\mathcal TP$s, however, we also need to compute $\beta$ -- a data-dependent term that scales the covariance matrix as in Eq.~\ref{eq:3}. To update $\beta$, we introduce recurrent expressions for the auxiliary variables:
    \begin{equation*}
        \begin{split}
            &\tilde z_i = z_i - \mu \\
            &a_n = \frac{v + \rho(n-2)}{(v-\rho)(v+\rho(n-1))}, \quad b_n = \frac{-\rho}{(v-\rho)(v+\rho(n-1))} \\
            &\beta_{n+1} = \beta_{n} + (a_n - b_n) \tilde z_n^2 + b_n (\sum_{i=1}^{n} \tilde z_i)^2 - b_{n-1} (\sum_{i=1}^{n-1} \tilde z_i)^2.
        \end{split}
    \end{equation*}
    From these equations, we see that computational complexity of making predictions in exchangeable $\mathcal GP$s or $\mathcal TP$s scales linearly with the number of observations, i.e. $\mathcal O(n)$ instead of a general $\mathcal O(n^3)$ case where one needs to compute an inverse covariance matrix.

    So far, we have constructed an exchangeable Student-t process in the latent space $\mathcal Z$. By coupling it with a bijective Real NVP mapping, we get an exchangeable process in space $\mathcal X$. Although we do not have an explicit analytic form of the transitions in $\mathcal X$, we still can sample from this process and evaluate the predictive distribution via the change of variables formula in Eq.~\ref{eq:4}.

    \subsection{Training}

    Having an easy-to-evaluate autoregressive distribution $p(\bm x_{n+1}| \bm x_{1:n})$ allows us to use a training scheme that is common for RNNs, i.e. maximise the likelihood of the next element in the sequence at every step. Thus, our objective function for a single sequence of fixed length $N$ can be written as  $\mathcal L = \sum_{n=0}^{N-1} \log p(\bm x_{n+1} | \bm x_{1:n})$, which is equivalent to maximising the joint log-likelihood $\log p(\bm x_1, \dots, \bm x_N)$. While we do have a closed-form expression for the latter, we chose not to use it during training in order to minimize the difference between the implementation of training and testing phases. Note that at test time, dealing with the joint log-likelihood would be inconvenient or even impossible due to high memory costs when $N$ gets large, which again motivates the use of a recurrent formulation.

     During training, we update the weights of the Real NVP model and also learn the parameters of the prior Student-t distribution. For the latter, we have three trainable parameters per dimension: degrees of freedom $\nu^d$, variance $v^d$ and covariance $\rho^d$. The mean $\mu^d$ is fixed to 0 for every $d$ and is not updated during training.

    \section{Experiments}
    \label{sec:experiments}

    In this section, we will consider a few problems that fit naturally into the framework of modeling exchangeable data.
    We chose to work with sequences of images, so the results are easy to analyse; yet \name\space does not make any image-specific assumptions, and our conclusions can generalise to other types of data. Specifically, for non-image data, one can use a general-purpose Real NVP coupling layer as proposed by~\citet{papamakarios17}. In contrast to the original Real NVP model, which uses convolutional architecture for scaling and translation functions in Eq.~\ref{eq:9}, a general implementation has $s$ and $t$ composed from fully connected layers. We experimented with both convolutional and non-convolutional architectures, the details of which are given in Appendix C.

    In our experiments, the models are trained on image sequences of length 20. We form each sequence by uniformly sampling a class and then selecting 20 random images from that class. This scheme implies that a model is trained to implicitly infer a class label that is global to a sequence. In what follows, we will see how this property can be used in a few tasks.

    \subsection{Conditional image generation}
    \label{imgen}

    We first consider a problem of generating samples conditionally on a set of images, which reduces to sampling from a predictive distribution. This is different from a general Bayesian approach, where one needs to infer the posterior over some meaningful latent variable and then `decode' it.

    To draw samples from $p(\bm x_{n+1}|\bm x_{1:n})$, we first sample $\bm z \sim p(\bm z_{n+1}|\bm z_{1:n})$ and then compute the inverse Real NVP mapping: $\bm x = f^{-1}(\bm z)$. Since we assumed that dimensions of $\bm z$ are independent, we can sample each $z^d$ from a univariate Student-t distribution. To do so, we modified Bailey's polar t-distribution generation method~\cite{bailey94} to be computationally efficient for GPU. Its algorithm is given in Appendix D.

    In Figure~\ref{fig:1}, we show samples from the prior distribution $p(\bm x_1)$ and conditional samples from a predictive distribution $p(\bm x_{n+1}|\bm x_{1:n})$ at steps $n=1,\dots,20$. Here, we used a convolutional Real NVP model as a part of \name. The model was trained on Omniglot~\cite{lake15} same-class image sequences of length 20 and we used the train-test split and preprocessing as defined by~\citet{vinyals16}. Namely, we resized the images to $28 \times 28$ pixels and augmented the dataset with rotations by multiples of 90 degrees yielding 4,800 and 1,692 classes for training and testing respectively.
    \begin{figure}[htb]
        \centering
        \includegraphics[width=0.7\linewidth]{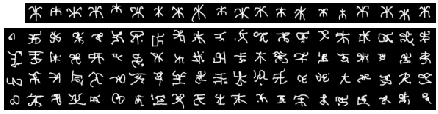}

        \caption{Samples generated conditionally on the sequence of the unseen Omniglot character class. An input sequence is shown in the top row and samples in the bottom 4 rows. Every column of the bottom subplot contains 4 samples from the predictive distribution conditioned on the input images up to and including that column. That is, the 1st column shows samples from the prior $p(\bm x)$ when no input image is given; the 2nd column shows samples from $p(\bm x| \bm x_1)$ where $\bm x_1$ is the 1st input image in the top row and so on.}

        \label{fig:1}
    \end{figure}%

    To better understand how \name~ behaves, we test it on special types of input sequences that were not seen during training. In Appendix E, we give an example where the same image is used throughout the sequence. In that case, the variability of the samples reduces as the models gets more of the same input. This property does not hold for the neural statistician model~\cite{edwards17}, discussed in Section~\ref{related_work}. As mentioned earlier, the neural statistician computes the approximate posterior $q(\bm c|\bm x_1,\dots,\bm x_n)$  and then uses its mean to sample $\bm x$ from a conditional model  $p(\bm x|\bm c_{mean})$. This scheme does not account for the variability in the inputs as a consequence of applying mean pooling over the features of $\bm x_1,\dots,\bm x_n$ when computing $q(\bm c|\bm x_1,\dots,\bm x_n)$. Thus, when all $x_i$'s are the same, it would still sample different instances from the class specified by $x_i$. Given the code provided by the authors of the neural statistician and following an email exchange, we could not reproduce the results from their paper, so we refrained from making any direct comparisons.

    More generated samples from convolutional and non-convolutional architectures trained on MNIST~\cite{mnist}, Fashion-MNIST~\cite{xiao17} and CIFAR-10~\cite{cifar10} are given in the appendix. For a couple of these models, we analyse the parameters of the learnt latent distributions (see Appendix F).

    \subsection{Few-shot learning}

    Previously, we saw that \name~ can generate images of the unseen classes even after being conditioned on a couple of examples. In this section, we will see how one can use its conditional probabilities not only for generation, but also for a few-shot classification.

    We evaluate the few-shot learning accuracy of the model from Section~\ref{imgen} on the unseen Omniglot characters from the 1,692 testing classes following the $n$-shot and $k$-way classification setup proposed by~\citet{vinyals16}. For every test case, we randomly draw a test image $\bm x_{n+1}$ and a sequence of $n$ images from the target class. At the same time, we draw $n$ images for every of the $k-1$ random decoy classes. To classify an image $\bm x_{n+1}$, we compute ${p(\bm x_{n+1}|\bm x^{C=i}_{1:n})}$ for each class ${i = 1\dots k}$ in the batch. An image is classified correctly when the conditional probability is highest for the target class compared to the decoy classes. This evaluation is performed 20 times for each of the test classes and the average classification accuracy is reported in Table~\ref{table:1}.

    For comparison, we considered three models from~\citet{vinyals16}: \textbf{(a)} k-nearest neighbours (k-NN), where matching is done on raw pixels (Pixels), \textbf{(b)} k-NN with matching on discriminative features from a state-of-the-art classifier (Baseline Classifier), and \textbf{(c)} Matching networks.

    We observe that \name~ model from Section~\ref{imgen} outperforms the baseline classifier, despite having been trained on relatively long sequences with a generative objective, i.e.  maximising the likelihood of the input images.  Yet, it cannot compete with matching networks -- a model tailored for a few-shot learning and trained in a discriminative way on short sequences such that its test-time protocol exactly matches the training time protocol. One can argue, however, that a comparison between models trained generatively and discriminatively is not fair. Generative modelling is a more general, harder problem to solve than discrimination, so a generatively trained model may waste a lot of statistical power on modelling aspects of the data which are irrelevant for the classification task. To verify our intuition, we fine-tuned \name~ with a discriminative objective, i.e. maximising the likelihood of correct labels in $n$-shot, $k$-way classification episodes formed from the training examples of Omniglot. While we could sample a different $n$ and $k$ for every training episode like in matching networks, we found it sufficient to fix $n$ and $k$ during training. Namely, we chose the setting with $n=1$ and $k=20$. From Table~\ref{table:1}, we see that this additional discriminative training makes \name~ competitive with state-of-the-art models across all $n$-shot and $k$-way tasks.

    \begin{table}[tbh]
        \small
        \caption{Classification accuracy for a few-shot learning task on the Omniglot dataset.}
        \label{table:1}
        \begin{center}
            \begin{tabular}{l@{\hskip \colspaceL}r@{\hskip \colspaceS}r@{\hskip \colspaceL}r@{\hskip \colspaceS}r@{\hskip \colspaceS}}
                \toprule
                \multirow{2}{*}{\b{Model}} & \multicolumn{2}{c}{\b{5-way}} &  \multicolumn{2}{c}{\b{20-way}}\\
                ~ &1-shot & 5-shot & 1-shot & 5-shot \\
                \midrule
                \b{\abbr{Pixels}~\cite{vinyals16}} & \t{41.7\%} & \t{63.2\%} & \t{26.7\%} & \t{42.6\%} \\
                \b{\abbr{Baseline Classifier}~\cite{vinyals16}} & \t{80.0\%} & \t{95.0\%} & \t{69.5\%} & \t{89.1\%} \\
                \b{\abbr{Matching Nets}~\cite{vinyals16}} & \t{98.1\%} & \t{98.9\%} & \t{93.8\%} & \t{98.5\%} \\
                \midrule
                \b{\abbr{BRUNO}} & \t{86.3\%} & \t{95.6\%} & \t{69.2\%} & \t{87.7\%} \\
                \b{\abbr{BRUNO ($\textnormal{discriminative fine-tuning}$)}} & \t{97.1\%} & \t{99.4\%} & \t{91.3\%} & \t{97.8\%} \\
                \bottomrule
            \end{tabular}
        \end{center}
    \end{table}

    As an extension to the few-shot learning task, we showed that \name~ could also be used for online set anomaly detection. These experiments can be found in Appendix H.

    \subsection{$\mathcal{GP}$-based models}
    \label{section:gp}
    In practice, we noticed that training $\mathcal{TP}$-based models can be easier compared to $\mathcal{GP}$-based models as they are more robust to anomalous training inputs and are less sensitive to the choise of hyperparameters. Under certain conditions, we were not able to obtain convergent training with $\mathcal{GP}$-based models which was not the case when using $\mathcal{TP}$s; an example is given in Appendix G. However, we found a few heuristics that make for a successful training such that $\mathcal{TP}$ and $\mathcal{GP}$-based models perform equally well in terms of test likelihoods, sample quality and few-shot classification results. For instance, it was crucial to use weight normalisation with a data-dependent initialisation of parameters of the Real NVP~\cite{salimans16}. As a result, one can opt for using $\mathcal {GP}$s due to their simpler implementation. Nevertheless, a Student-t process remains a strictly richer model class for the latent space with negligible additional computational costs.

    \section{Discussion and conclusion}
    \label{sec:discussion}

    In this paper, we introduced \name, a new technique combining deep learning and Student-t or Gaussian processes for modelling exchangeable data. With this architecture, we may carry out implicit Bayesian inference, avoiding the need to compute posteriors and eliminating the high computational cost or approximation errors often associated with explicit Bayesian inference.

    Based on our experiments, \name\space shows promise for applications such as conditional image generation, few-shot concept learning, few-shot classification and online anomaly detection. The probabilistic construction makes the \name\ approach particularly useful and versatile in transfer learning and multi-task situations. To demonstrate this, we showed that \name\ trained in a generative way achieves good performance in a downstream few-shot classification task without any task-specific retraining. Though, the performance can be significantly improved with discriminative fine-tuning.

    Training \name\ is a form of meta-learning or learning-to-learn: it learns to perform Bayesian inference on various sets of data. Just as encoding translational invariance in convolutional neural networks seems to be the key to success in vision applications, we believe that the notion of exchangeability is equally central to data-efficient meta-learning. In this sense, architectures like \name\ and Deep Sets~\cite{zaheer17} can be seen as the most natural starting point for these applications.

    As a consequence of exchangeability-by-design, \name\ is endowed with a hidden state which integrates information about all inputs regardless of sequence length. This desired property for meta-learning is usually difficult to ensure in general RNNs as they do not automatically generalise to longer sequences than they were trained on and are sensitive to the ordering of inputs. Based on this observation, the most promising applications for \name\ may fall in the many-shot meta-learning regime, where larger sets of data are available in each episode. Such problems naturally arise in privacy-preserving on-device machine learning, or federated meta-learning~\cite{chen2018federated}, which is a potential future application area for \name.

    \section*{Acknowledgements}
    We would like to thank Lucas Theis for his conceptual contributions to BRUNO, Conrado Miranda and Frederic Godin for their helpful comments on the paper, Wittawat Jitkrittum for useful discussions, and Lionel Pigou for setting up the hardware.

    \bibliography{main}

\begin{thebibliography}{}

\bibitem[Aldous et~al., 1985]{aldous}
Aldous, D., Hennequin, P., Ibragimov, I., and Jacod, J. (1985).
\newblock {\em Ecole d'Ete de Probabilites de Saint-Flour XIII, 1983}.
\newblock Lecture Notes in Mathematics. Springer Berlin Heidelberg.

\bibitem[Bailey, 1994]{bailey94}
Bailey, R.~W. (1994).
\newblock Polar generation of random variates with the {$t$}-distribution.
\newblock {\em Math. Comp.}, 62(206):779--781.

\bibitem[Chen et~al., 2018]{chen2018federated}
Chen, F., Dong, Z., Li, Z., and He, X. (2018).
\newblock Federated meta-learning for recommendation.
\newblock {\em arXiv preprint arXiv:1802.07876}.

\bibitem[Clevert et~al., 2016]{clevert16}
Clevert, D., Unterthiner, T., and Hochreiter, S. (2016).
\newblock Fast and accurate deep network learning by exponential linear units
  ({ELU}s).
\newblock In {\em Proceedings of the 4th International Conference on Learning
  Representations}.

\bibitem[Dinh et~al., 2014]{dinh14}
Dinh, L., Krueger, D., and Bengio, Y. (2014).
\newblock {NICE:} non-linear independent components estimation.
\newblock {\em arXiv preprint}, abs/1410.8516.

\bibitem[Dinh et~al., 2017]{dinh17}
Dinh, L., Sohl-Dickstein, J., and Bengio, S. (2017).
\newblock Density estimation using {R}eal {NVP}.
\newblock In {\em Proceedings of the 5th International Conference on Learning
  Representations}.

\bibitem[Edwards and Storkey, 2017]{edwards17}
Edwards, H. and Storkey, A. (2017).
\newblock Towards a neural statistician.
\newblock In {\em Proceedings of the 5th International Conference on Learning
  Representations}.

\bibitem[Ghahramani and Heller, 2006]{ghahramani06}
Ghahramani, Z. and Heller, K.~A. (2006).
\newblock Bayesian sets.
\newblock In Weiss, Y., Sch\"{o}lkopf, B., and Platt, J.~C., editors, {\em
  Advances in Neural Information Processing Systems 18}, pages 435--442. MIT
  Press.

\bibitem[Heller and Ghahramani, 2006]{heller06}
Heller, K.~A. and Ghahramani, Z. (2006).
\newblock A simple bayesian framework for content-based image retrieval.
\newblock In {\em {IEEE} Computer Society Conference on Computer Vision and
  Pattern Recognition}, pages 2110--2117.

\bibitem[Kingma and Welling, 2014]{kingma14b}
Kingma, D.~P. and Welling, M. (2014).
\newblock Auto-encoding variational bayes.
\newblock In {\em Proceedings of the 2nd International Conference on Learning
  Representations}.

\bibitem[Krizhevsky, 2009]{cifar10}
Krizhevsky, A. (2009).
\newblock Learning multiple layers of features from tiny images.
\newblock Technical report.

\bibitem[Lake et~al., 2015]{lake15}
Lake, B.~M., Salakhutdinov, R., and Tenenbaum, J.~B. (2015).
\newblock Human-level concept learning through probabilistic program induction.
\newblock {\em Science}.

\bibitem[LeCun et~al., 1998]{mnist}
LeCun, Y., Cortes, C., and Burges, C.~J. (1998).
\newblock The {MNIST} database of handwritten digits.

\bibitem[Papamakarios et~al., 2017]{papamakarios17}
Papamakarios, G., Murray, I., and Pavlakou, T. (2017).
\newblock Masked autoregressive flow for density estimation.
\newblock In {\em Advances in Neural Information Processing Systems 30}, pages
  2335--2344.

\bibitem[Rasmussen and Williams, 2005]{rasmussen05}
Rasmussen, C.~E. and Williams, C. K.~I. (2005).
\newblock {\em Gaussian Processes for Machine Learning (Adaptive Computation
  and Machine Learning)}.
\newblock The MIT Press.

\bibitem[Rezende and Mohamed, 2015]{rezende15}
Rezende, D. and Mohamed, S. (2015).
\newblock Variational inference with normalizing flows.
\newblock In {\em Proceedings of the 32nd International Conference on Machine
  Learning}, volume~37 of {\em Proceedings of Machine Learning Research}, pages
  1530--1538.

\bibitem[Rezende et~al., 2014]{rezende14}
Rezende, D.~J., Mohamed, S., and Wierstra, D. (2014).
\newblock Stochastic backpropagation and approximate inference in deep
  generative models.
\newblock In {\em Proceedings of the 31st International Conference on Machine
  Learning}, pages 1278--1286.

\bibitem[Salimans and Kingma, 2016]{salimans16}
Salimans, T. and Kingma, D.~P. (2016).
\newblock Weight normalization: A simple reparameterization to accelerate
  training of deep neural networks.
\newblock In {\em Proceedings of the 30th International Conference on Neural
  Information Processing Systems}.

\bibitem[Shah et~al., 2014]{shah14}
Shah, A., Wilson, A.~G., and Ghahramani, Z. (2014).
\newblock Student-t processes as alternatives to gaussian processes.
\newblock In {\em Proceedings of the 17th International Conference on
  Artificial Intelligence and Statistics}, pages 877--885.

\bibitem[Szabo et~al., 2016]{SzaSriPocGre16}
Szabo, Z., Sriperumbudur, B., Poczos, B., and Gretton, A. (2016).
\newblock Learning theory for distribution regression.
\newblock {\em Journal of Machine Learning Research}, 17(152).

\bibitem[Theis et~al., 2016]{theis16}
Theis, L., van~den Oord, A., and Bethge, M. (2016).
\newblock A note on the evaluation of generative models.
\newblock In {\em Proceedings of the 4th International Conference on Learning
  Representations}.

\bibitem[Tieleman and Hinton, 2012]{rmsprop}
Tieleman, T. and Hinton, G. (2012).
\newblock {Lecture 6.5 - {RmsProp}: Divide the gradient by a running average of
  its recent magnitude}.
\newblock COURSERA: Neural Networks for Machine Learning.

\bibitem[Vinyals et~al., 2016a]{vinyals16b}
Vinyals, O., Bengio, S., and Kudlur, M. (2016a).
\newblock Order matters: Sequence to sequence for sets.
\newblock In {\em Proceedings of the 4th International Conference on Learning
  Representations}.

\bibitem[Vinyals et~al., 2016b]{vinyals16}
Vinyals, O., Blundell, C., Lillicrap, T., Kavukcuoglu, K., and Wierstra, D.
  (2016b).
\newblock Matching networks for one shot learning.
\newblock In {\em Advances in Neural Information Processing Systems 29}, pages
  3630--3638.

\bibitem[Xiao et~al., 2017]{xiao17}
Xiao, H., Rasul, K., and Vollgraf, R. (2017).
\newblock Fashion-mnist: a novel image dataset for benchmarking machine
  learning algorithms.
\newblock {\em arXiv preprint}, abs/1708.07747.

\bibitem[Zaheer et~al., 2017]{zaheer17}
Zaheer, M., Kottur, S., Ravanbakhsh, S., Poczos, B., Salakhutdinov, R.~R., and
  Smola, A.~J. (2017).
\newblock Deep sets.
\newblock In {\em Advances in Neural Information Processing Systems 30}, pages
  3394--3404.

\end{thebibliography}
    \bibliographystyle{apalike}

\clearpage
 \appendix
  \section{Proofs}

    \subsection*{Lemma 1}
    \textit{Given an exchangeable sequence $(x_1, x_2, \dots, x_n)$ of random variables $x_i \in \mathcal X$ and a bijective mapping $f: \mathcal X \mapsto \mathcal Z$, the sequence $(f(x_1), f(x_2), \dots, f(x_n))$ is exchangeable.}

    %\raggedright
    \textbf{\textit{Proof.}} Consider a vector function $\bm g: \mathbb R^n \mapsto \mathbb R^n$ such that \\
    ${(x_1, \dots, x_n) \mapsto (z_1=f(x_1), \dots, z_n=f(x_n))}$. A change of variable formula gives:
    \begin{equation*}
        p(x_1, x_2, \dots, x_n) = p(z_1, z_2, \dots, z_n) \left| \det \bm J \right|,
    \end{equation*}
    where $\det \bm J= \prod_{i=1}^n \frac{\partial f(x_i)}{ \partial x_i}$ is the determinant of the Jacobian of $\bm g$.
    Since both the joint probability of  $(x_1, x_2, \dots, x_n)$ and the $\left| \det \bm J \right|$ are invariant to the permutation of sequence entries, so must be $p(z_1, z_2, \dots, z_n)$. This proves that $(z_1, z_2, \dots, z_n)$ is exchangeable. $\square$

     \subsection*{Lemma 2}
\textit{Given two exchangeable sequence $\bm{x} = (x_1, x_2, \dots, x_n)$ and $\bm{y} = (y_1, y_2, \dots, y_n)$ of random variables, where $x_i$ is independent from $y_j$ for $\forall i,j$, the concatenated sequence $\bm{x} ^\frown \bm{y} = ((x_1, y_1), (x_2, y_2), \dots, (x_n, y_n))$ is exchangeable as well.}

\textbf{\textit{Proof.}} For any permutation $\pi$, as both sequences $\bm x$ and $\bm y$ are exchangeable we have:
\begin{equation*}
\begin{aligned}
  p(x_1, x_2, \dots, x_n) p(y_1, y_2, \dots, y_n) = p(x_{\pi(1)}, x_{\pi(2)}, \dots, x_{\pi(n)}) p(y_{\pi(1)}, y_{\pi(2)}, \dots, y_{\pi(n)}).
  \end{aligned}
\end{equation*}
Independence between elements in $\bm x$ and $\bm y$ allows to write it as a joint distribution:
\begin{equation*}
\begin{aligned}
  p((x_1, y_1), (x_2, y_2) \dots, (x_n, y_n)) = p((x_{\pi(1)},y_{\pi(1)}), (x_{\pi(2)},y_{\pi(2)}), \dots, (x_{\pi(n)},y_{\pi(n)})),
  \end{aligned}
\end{equation*}
and thus the sequence $\bm{x} ^\frown \bm{y}$ is exchangeable. $\square$

This Lemma justifies our construction with $D$ independent exchangeable processes in the latent space as given in A1 from Section 3.3.

    \section{Derivation of recurrent Bayesian updates for exchangeable Student-t and Gaussian processes}

    We assume that $\bm x = (x_1, x_2, \dots x_n) \in \mathbb{R}^n$ follows a multivariate Student-t distribution $MVT_n(\nu, \bm \mu, \bm K)$ with
    degrees of freedom $\nu \in \mathbb{R}_+ \setminus[0,2] $, mean $\bm \mu \in \mathbb{R}^n$ and a positive definite $n \times n$ covariance matrix $\bm K$. Its density is given by:
    \begin{equation}
        \begin{split}
            p(\bm x) = \frac{\Gamma(\frac{\nu+n}{2})}{((\nu-2)\pi)^{n/2}\Gamma(\nu/2)}| \bm K|^{-1/2} \bigg(1+\frac{(\bm x- \bm \mu)^T \bm K^{-1}( \bm x- \bm \mu)}{\nu-2} \bigg) ^{-\frac{\nu+n}{2}}.
        \end{split}
    \end{equation}%
    Note that this parameterization of the multivariate t-distribution as defined by \citet{shah14} is slightly different from the commonly used one. We used this parametrization as it makes the formulas simpler.

    If we partition $\bm x$ into two consecutive parts ${\bm x_a \in \mathbb{R}^{n_a}}$ and ${\bm x_b \in \mathbb{R}^{n_b}}$:
    \begin{equation*}
        \begin{bmatrix}
            \bm x_a \cr \bm x_b
        \end{bmatrix} \sim MVT_n\Bigg(\nu, \begin{bmatrix}
                                               \bm \mu_a \cr \bm \mu_b
        \end{bmatrix},
        \begin{bmatrix}
            \bm K_{aa} & \bm K_{ab} \cr
            \bm K_{ba} & \bm K_{bb}
        \end{bmatrix}\Bigg),
    \end{equation*}
    the conditional distribution $p(\bm x_b| \bm x_a)$ is given by:
    \begin{equation}
        \label{eq:0}
        p(\bm x_b| \bm x_a) = MVT_{n_b}(\nu + n_a, \bm{\tilde{\mu}_b}, \frac{\nu + \beta_a -2}{\nu + n_a-2} \bm{\tilde K}_{bb}),
    \end{equation}
    where
    \begin{equation*}
        \begin{split}
            &\bm{\tilde{\mu}_b} = \bm K_{ba}\bm K_{aa}^{-1}(\bm x_a - \bm \mu_a) + \bm \mu_b\\
            &\beta_a = (\bm x_a-\bm \mu_a)^T \bm K_{aa}^{-1} (\bm x_a - \bm \mu_a)\\
            &\bm{\tilde K}_{bb} = \bm K_{bb} - \bm K_{ba} \bm K_{aa}^{-1} \bm K_{ab}.
        \end{split}
    \end{equation*}

    Derivation of this result is given in the appendix of~\cite{shah14}. Let us now simplify these equations for the case
    of exchangeable sequences with the following covariance structure:
    \begin{equation*}
        \bm K = \begin{pmatrix}
                    v & \rho &\cdots & \rho \cr
                    \rho & v &\cdots & \rho \cr
                    \vdots  & \vdots  & \ddots & \vdots \cr
                    \rho & \rho & \cdots & v
        \end{pmatrix}.
    \end{equation*}

    In our problem, we are interested in doing one-step predictions, i.e. computing a univariate density $p(x_{n+1}|x_{1:n})$ with parameters $\nu_{n+1}$, $\mu_{n+1}$, $v_{n+1}$. Therefore, in Eq.~\ref{eq:0} we can take: $n_b=1$, $n_a=n$, $\bm x_a = x_{1:n} \in \mathbb R^{n}$, $ \bm x_b = x_{n+1} \in \mathbb R$,  $\bm K_{aa} = \bm K_{1:n, 1:n}$,  $\bm K_{ab} = \bm K_{1:n, n+1}$, ${\bm K_{ba} = \bm K_{n+1, 1:n}}$ and ${\bm K_{bb} = \bm K_{n+1, n+1} = v}$.

    Computing the parameters of the predictive distribution requires the inverse of $\bm K_{aa}$, which we can find using the Sherman-Morrison formula:
    \begin{equation*}
        \bm{K}_{aa}^{-1} = (\bm A+ \bm u \bm v^{T})^{-1}=\bm A^{-1}-\frac{\bm A^{-1}\bm u \bm v^{T}\bm A^{-1}}{1+\bm v^{T}\bm A^{-1}\bm u},
    \end{equation*}
    with
    \begin{equation*}
        \bm A = \begin{pmatrix}
                    v - \rho & 0 &\cdots & 0\cr
                    0 & v - \rho &\cdots & 0 \cr
                    \vdots  & \vdots  & \ddots & \vdots \cr
                    0 & 0 & \cdots & v - \rho
        \end{pmatrix}, \,
    \end{equation*}
    \begin{equation*}
        \bm u = \begin{pmatrix}
                    \rho \cr \rho \cr \vdots \cr \rho
        \end{pmatrix},\,
        \bm v = \begin{pmatrix}
                    1 \cr 1 \cr \vdots \cr 1
        \end{pmatrix}.
    \end{equation*}

    After a few steps, the inverse of $\bm K_{aa}$ is:
    \begin{equation*}
        \bm K_{aa}^{-1} = \begin{pmatrix}
                              a_n & b_n &\cdots & b_n \cr
                              b_n & a_n &\cdots & b_n \cr
                              \vdots  & \vdots  & \ddots & \vdots \cr
                              b_n & b_n & \cdots & a_n
        \end{pmatrix}
    \end{equation*} with
    \begin{align*}
        a_n = \frac{v + \rho (n-2)}{(v-\rho)(v + \rho(n-1))},\\
        b_n = \frac{-\rho}{(v-\rho)(v+\rho (n-1))}. \\
    \end{align*}%
    Note that entries of $\bm K_{aa}^{-1}$ explicitly depend on $n$.

    Equations for the mean and variance of the predictive distribution require the following term:
    \begin{equation*}
        \bm K_{ba} \bm K_{aa}^{-1} = \begin{pmatrix}
                                         \rho & \rho & \cdots & \rho
        \end{pmatrix} \bm K_{aa}^{-1}
        = \Big\{ \frac{\rho}{v + \rho(n-1)}\Big\}_{1:n},
    \end{equation*}
    which is a $1 \times n$ vector.

    With this in mind, it is easy to derive the following recurrence:
    \begin{equation*}
        d_{n} = \frac{\rho}{v+\rho(n-1)}
    \end{equation*}
    \begin{equation*}
        \mu_{n+1} = (1-d_n) \mu_n + d_n x_n
    \end{equation*}
    \begin{equation*}
        v_{n+1} = (1-d_n) v_n + d_n (\rho-v).
    \end{equation*}

    Finally, let us derive recurrent equations for ${\beta_{n+1} = (\bm x_a - \bm \mu_a)^T K_{aa}^{-1} (\bm x_a - \bm \mu_a)}$.

    Let $\bm {\tilde x} = \bm x_a - \bm \mu_a$, then:
    \begin{equation*}
        \begin{aligned}
            &\beta_{n+1} = \bm {\tilde x} ^T K_{aa}^{-1} \bm{\tilde x} \\
            &= (a_n \tilde x_1 + b_n \sum_{i\neq 1}^{n} \tilde x_i, a_n \tilde x_2 + b_n \sum_{i\neq 2}^{n} \tilde x_i, \dots,
            a_n \tilde x_n + b_n \sum_{i\neq n} \tilde x_i)^T (\tilde x_1, \tilde x_2, \dots \tilde  x_n) \\
            &= (a_n - b_n) \sum_{i=1}^n \tilde x_i^2 + b_n (\sum_{i=1}^n \tilde x_i)^2.
        \end{aligned}
    \end{equation*}

    Similarly, $\beta_n$ from $p(x_{n}|x_{1:n-1})$ is:
    \begin{equation*}
        \beta_{n} = (a_{n-1} - b_{n-1}) \sum_{i=1}^{n-1} \tilde x_i^2 + b_{n-1} (\sum_{i=1}^{n-1} \tilde x_i)^2
    \end{equation*}
    \begin{equation*}
        \begin{aligned}
            &\beta_{n+1} = (a_n - b_n) (\sum_{i=1}^{n-1} \tilde x_i^2 + \tilde x_n^2 )+ b_n (\sum_{i=1}^{n}\tilde  x_i)^2 \\
            & = (a_n - b_n) \frac{\beta_n - b_{n-1}(\sum_{i=1}^{n-1} \tilde x_i)^2}{a_{n-1} - b_{n-1}} + (a_n - b_n)\tilde  x_n^2 + b_n (\sum_{i=1}^{n} \tilde x_i)^2.
        \end{aligned}
    \end{equation*}

    It is easy to show that $\frac{a_n-b_n}{a_{n-1}-b_{n-1}} = 1$, so $\beta_{n+1}$ can be written recursively as:
    \begin{equation*}
        \begin{aligned}
            &s_{n+1} = s_n + \tilde  x_n \\
            &\beta_{n+1} = \beta_n + (a_n - b_n) \tilde  x_n^2 + b_n (s_{n+1}^2 - s_{n}^2),
        \end{aligned}
    \end{equation*}
    with $s_1=0$.

    \section{Implementation details}

    For simple datasets, such as MNIST, we found it tolerable to use models that rely upon a general implementation of the Real NVP coupling layer similarly to \citet{papamakarios17}. Namely, when scaling and translation functions $s$ and $t$ are fully-connected neural networks. In our model, networks $s$ and $t$ share the parameters in the first two dense layers with 1024 hidden units and ELU nonlinearity~\cite{clevert16}. Their output layers are different: $s$ ends with a dense layer with $\tanh$ and $t$ ends with a dense layer without a nonlinearity. We stacked 6 coupling layers with alternating the indices of the transformed dimensions between odd and even as described by \citet{dinh14}. For the first layer, which implements a logit transformation of the inputs, namely ${f(x)=\textnormal{logit}(\alpha + (1 - 2 \alpha)x)}$, we used $\alpha=10^{-6}$. The logit transformation ensures that when taking the inverse mapping during sample generation, the outputs always lie within ${(\frac{-\alpha}{1-2\alpha}, \frac{1-\alpha}{1-2\alpha})}$.

    In Omniglot, Fashion MNIST and CIFAR-10 experiments, we built upon a Real NVP model originally designed for CIFAR-10 by \citet{dinh17}: a multi-scale architecture with deep convolutional residual networks in the coupling layers. Our main difference was the use of coupling layers with fully-connected $s$ and $t$ networks (as described above) placed on top of the original convolutional Real NVP model. We found that adding these layers allowed for a faster convergence and improved results. This is likely due to a better mixing of the information before the output of the Real NVP gets into the Student-t layer. We also found that using weight normalisation~\cite{salimans16} within every $s$ and $t$ function was crucial for successful training of large models.

    The model parameters were optimized using RMSProp~\cite{rmsprop} with a decaying learning rate starting from $10^{-3}$. Trainable parameters of a $\mathcal{TP}$ or $\mathcal{GP}$  were updated with a 10x smaller learning rate and were initialized as following: $\nu^d = 1000$, $v^d = 1.$, $\rho^d= 0.1$ for every dimension $d$. The mean $\mu^d$ was fixed at 0. For the Omniglot model, we used a batch size of 32,
  sequence length of 20 and trained for 200K iterations. The other models were trained for a smaller number of iterations, i.e. ranging from 50K to 100K updates.

  \newpage

    \section{Sampling from a Student-t distribution}
    \vskip -0.1in
    \begin{algorithm}
        \caption{Efficient sampling on GPU from a univariate t-distribution with mean $\mu$, variance $v$ and degrees of freedom $\nu$}
        \label{alg:1}
        \begin{algorithmic}
            \FUNCTION{sample($\mu, v, \nu$)}
            \STATE $a,b \gets \mathcal U(0,1)$
            \STATE $c \gets \min(a,b)$
            \STATE $r \gets \max(a,b)$
            \STATE $\alpha \gets \frac{2\pi c}{r}$
            \STATE $t \gets \cos(\alpha) \sqrt{(\nicefrac{\nu}{r^2})(r^{\nicefrac{-4}{\nu}} - 1)}$
            \STATE $\sigma \gets \sqrt{v\big(\frac{\nu-2}{\nu}\big)}$
            \STATE {\bfseries return} $\mu + \sigma t$
            \ENDFUNCTION
        \end{algorithmic}
    \end{algorithm}

    \section{Sample analysis}

    In Figure~\ref{fig:1}, which includes Figure 2 from the main text, we want to illustrate how sample variability depends on the variance of the inputs. From these examples, we see that in the case of a repeated input image, samples get more coherent as the number of conditioning inputs grows. It also shows that BRUNO does not merely generate samples according to the inferred class label.

    While Omngilot is limited to 20 images per class, we can experiment with longer sequences using CIFAR-10 or MNIST. In Figure~\ref{fig:1_2} and Figure~\ref{fig:1_3}, we show samples from the models trained on those datasets. In Figure~\ref{fig:1_4}, we also show more samples from the prior distribution $p(\bm x)$.

     \begin{figure}[h]
        \centering
        \subfigure{\includegraphics[width=0.49\linewidth]{img/sample_test_382_0.png}}
        \subfigure{\includegraphics[width=0.49\linewidth]{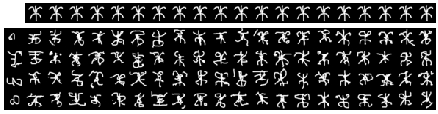}}
        \caption{Samples generated conditionally on images from an unseen Omniglot character class. \textit{Left:} input sequence of 20 images from one class. \textit{Right:} the same image is used as an input at every step.}
          \label{fig:1}
    \end{figure}%
    \begin{figure}[h]
        \centering
        \subfigure{\includegraphics[width=0.49\linewidth]{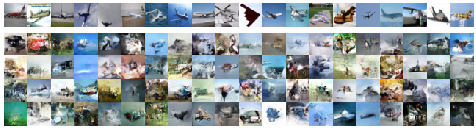}}
        \subfigure{\includegraphics[width=0.49\linewidth]{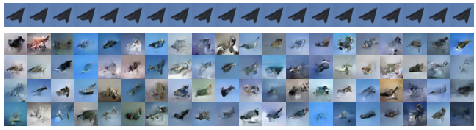}}
        \subfigure{\includegraphics[width=0.49\linewidth]{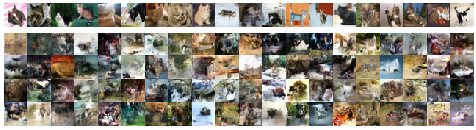}}
        \subfigure{\includegraphics[width=0.49\linewidth]{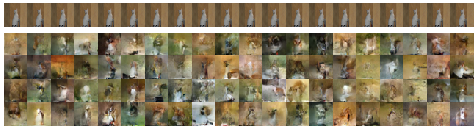}}
        \caption{CIFAR-10 samples from $p(\mathbf x | \mathbf x_{1:n})$ for every $n=480,\dots,500$. \textit{Left}: input sequence (given in the top row of each subplot) is composed of random same-class test images. \textit{Right}: same image is given as input at every step. In both cases, input images come from the test set of CIFAR-10 and the model was trained on all of the classes.}
          \label{fig:1_2}
    \end{figure}%
        \begin{figure}[ht]
        \centering
        \subfigure{\includegraphics[width=0.49\linewidth]{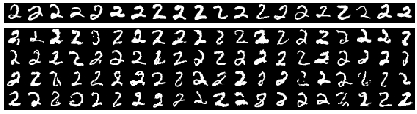}}
        \subfigure{\includegraphics[width=0.49\linewidth]{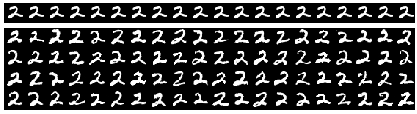}}
        \subfigure{\includegraphics[width=0.49\linewidth]{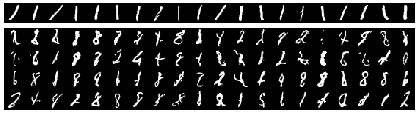}}
        \subfigure{\includegraphics[width=0.49\linewidth]{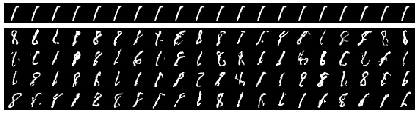}}
        \caption{MNIST samples from $p(\mathbf x | \mathbf x_{1:n})$ for every $n=480,\dots,500$. \textit{Left}: input sequence (given in the top row of each subplot) is composed of random same-class test images. \textit{Right}: same image is given as input at every step. In both cases, input images come from the test set of MNIST and the model was trained only on even digits, so it did not see digit `1' during training.}
          \label{fig:1_3}
    \end{figure}%

    \begin{figure}[ht]
        \centering
        \subfigure{\includegraphics[width=0.24\linewidth]{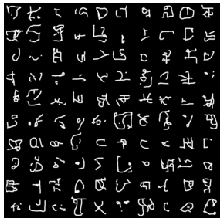}}
        \subfigure{\includegraphics[width=0.24\linewidth]{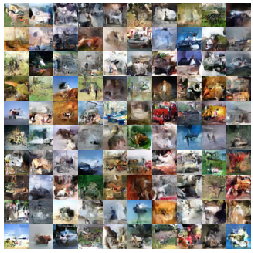}}
        \subfigure{\includegraphics[width=0.24\linewidth]{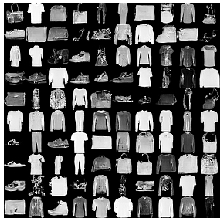}}
        \subfigure{\includegraphics[width=0.24\linewidth]{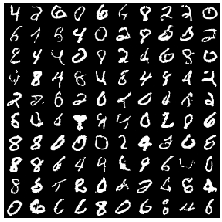}}
        \caption{Samples from the prior for the models trained on Omniglot, CIFAR-10, Fashion MNIST and MNIST (only trained on even digits).}
          \label{fig:1_4}
    \end{figure}%

 \newpage
 \section{Parameter analysis}

\begin{figure}[h]
        \subfigure{\includegraphics[width=0.32\linewidth]{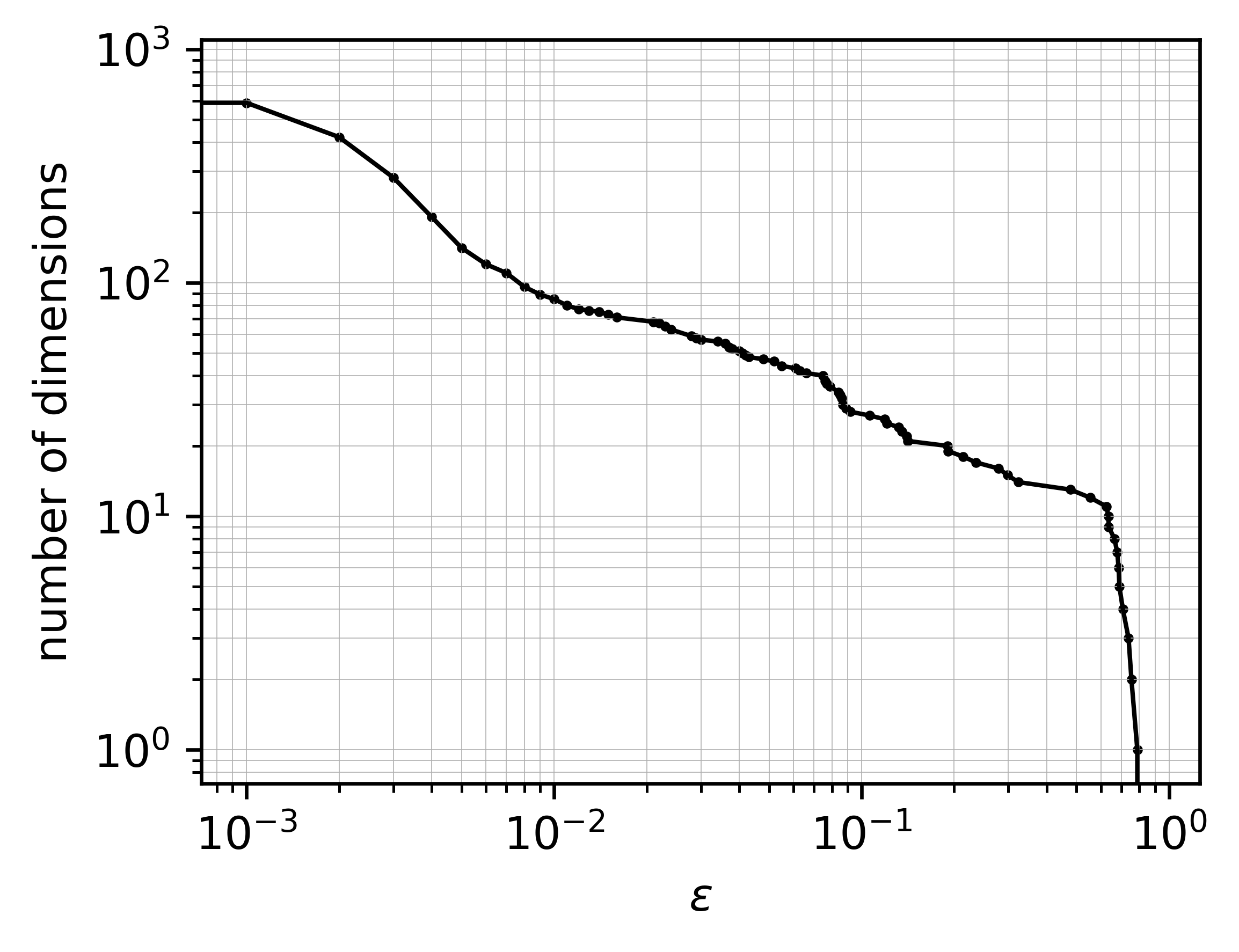}}
        \subfigure{\includegraphics[width=0.32\linewidth]{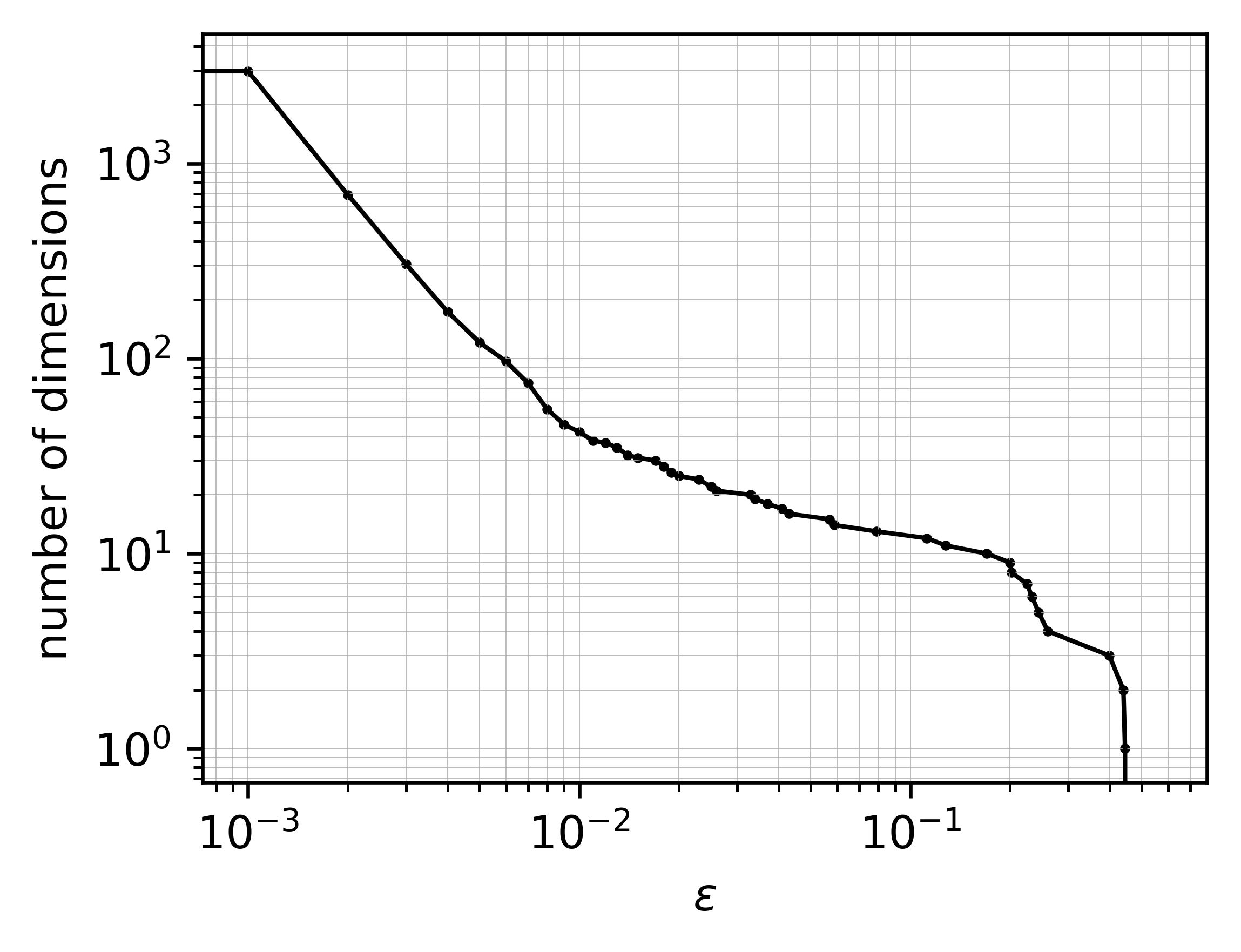}}
        \subfigure{\includegraphics[width=0.32\linewidth]{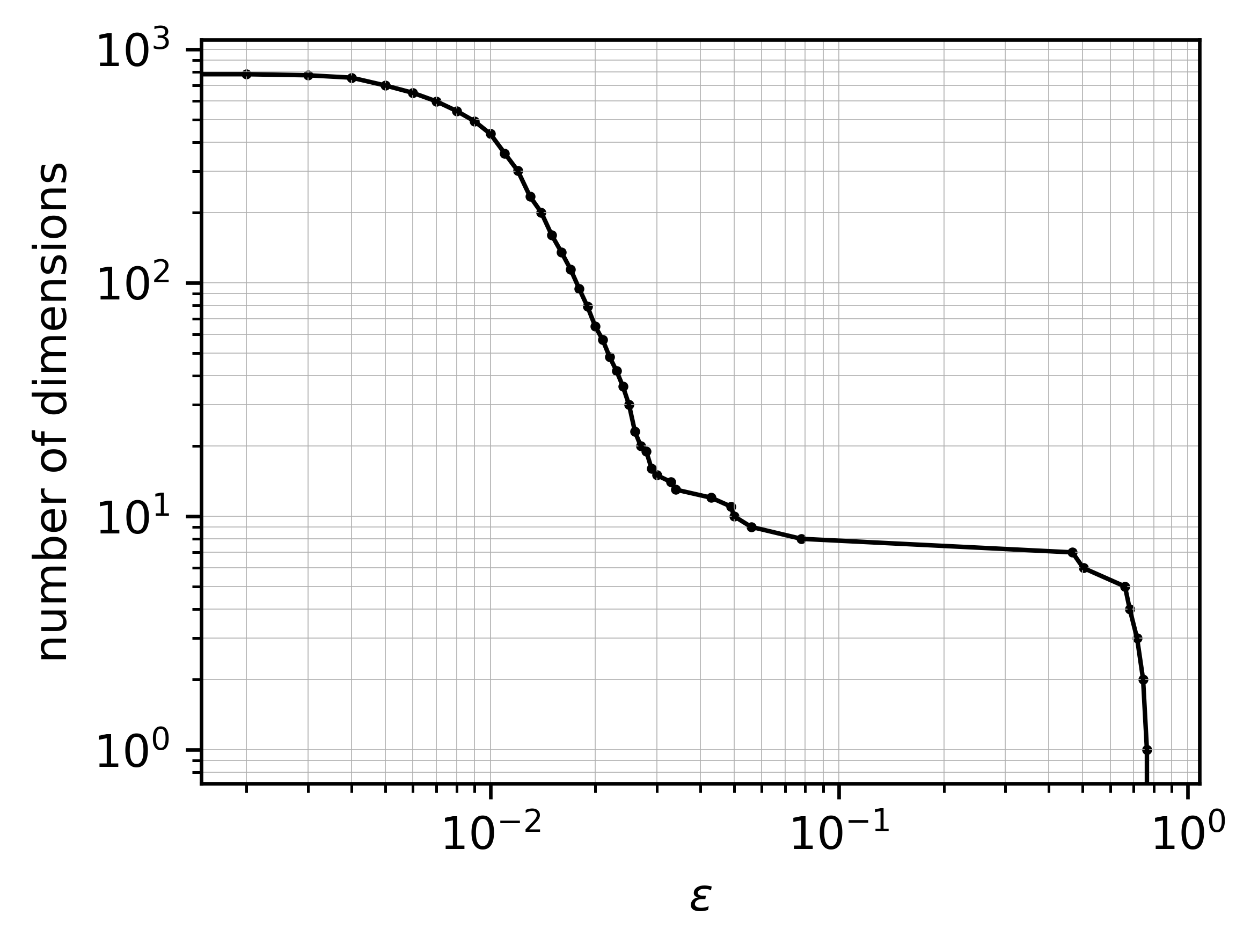}}
        \caption{Number of dimensions where $\nicefrac{\rho^d}{v^d} > \epsilon$ plotted on a double logarithmic scale.
        \textit{Left}: Omniglot model. \textit{Middle}: CIFAR-10 model \textit{Right}: Non-convolutional version of BRUNO trained on MNIST.}
        \label{fig:corr}
    \end{figure}

      After training a model, we observed that a majority of the processes in the latent space have low correlations $\nicefrac{\rho^d}{v^d}$, and thus their predictive distributions remain close to the prior. Figure~\ref{fig:corr} plots the number of dimensions where correlations exceed a  certain value on the x-axis. For instance, MNIST model has 8 dimensions where the correlation is higher than 0.1. While we have not verified it experimentally, it is reasonable to expect those dimensions to capture information about visual features of the digits.

\begin{figure}[h]
            \subfigure{\includegraphics[width=0.32\linewidth]{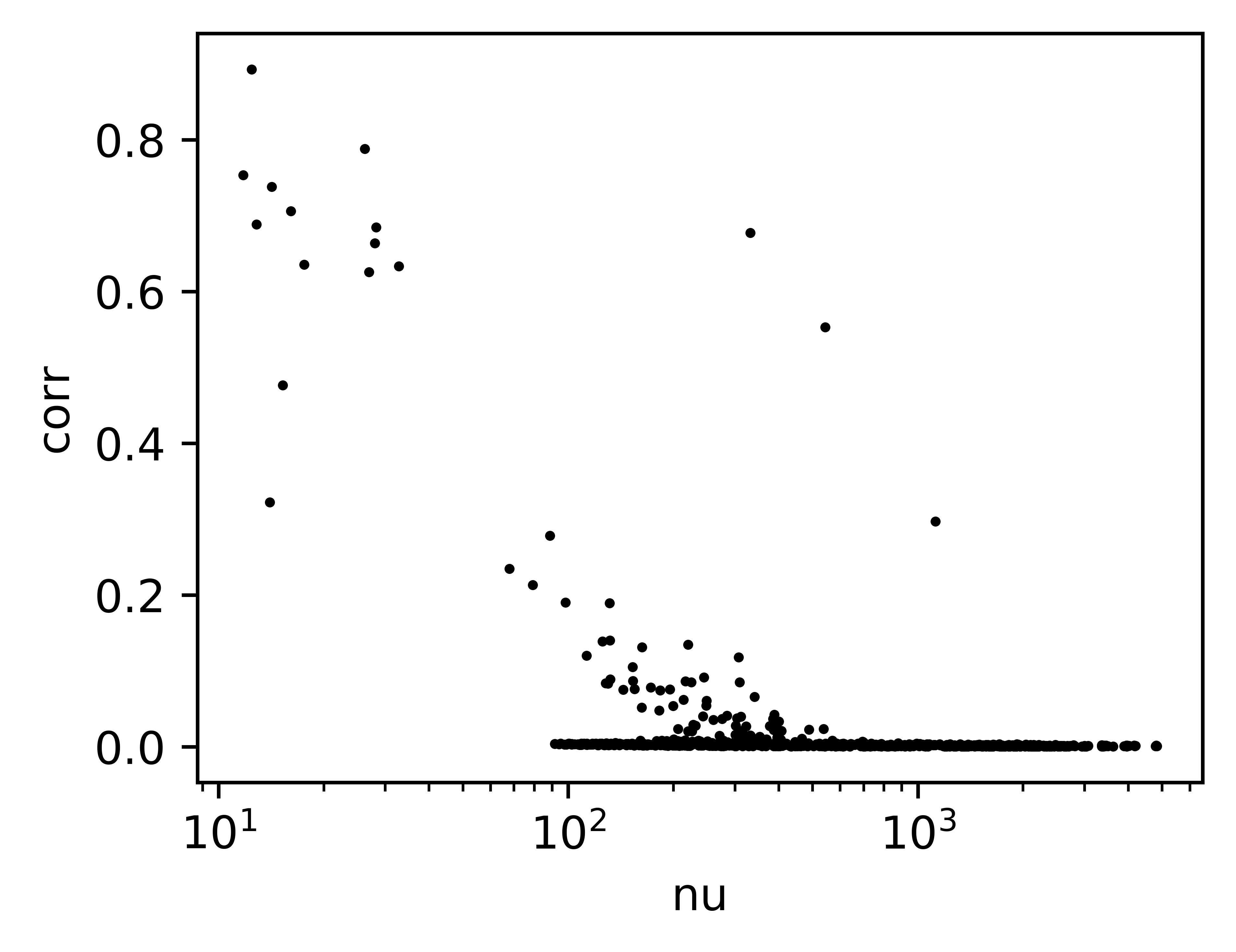}}
            \subfigure{\includegraphics[width=0.32\linewidth]{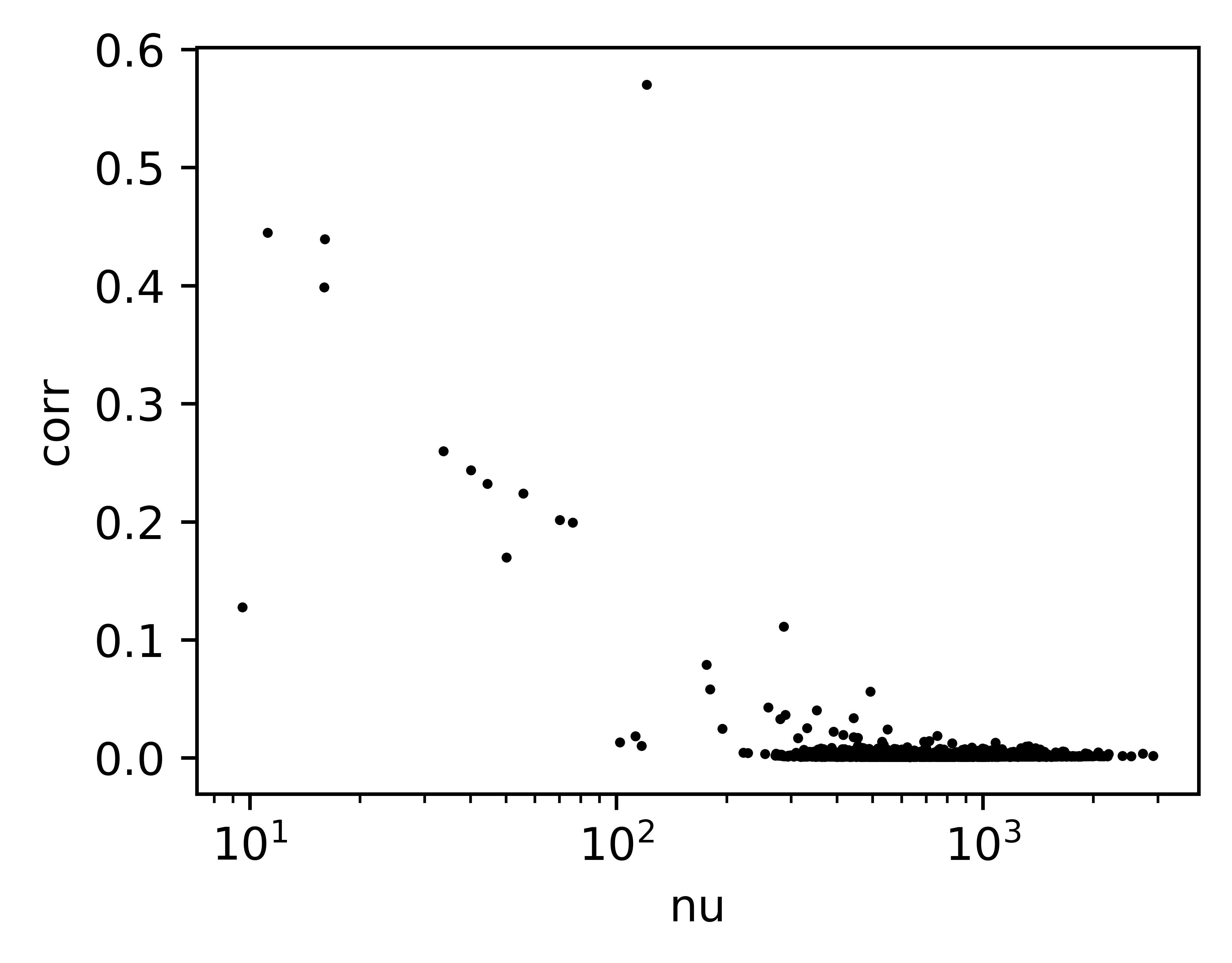}}
            \subfigure{\includegraphics[width=0.32\linewidth]{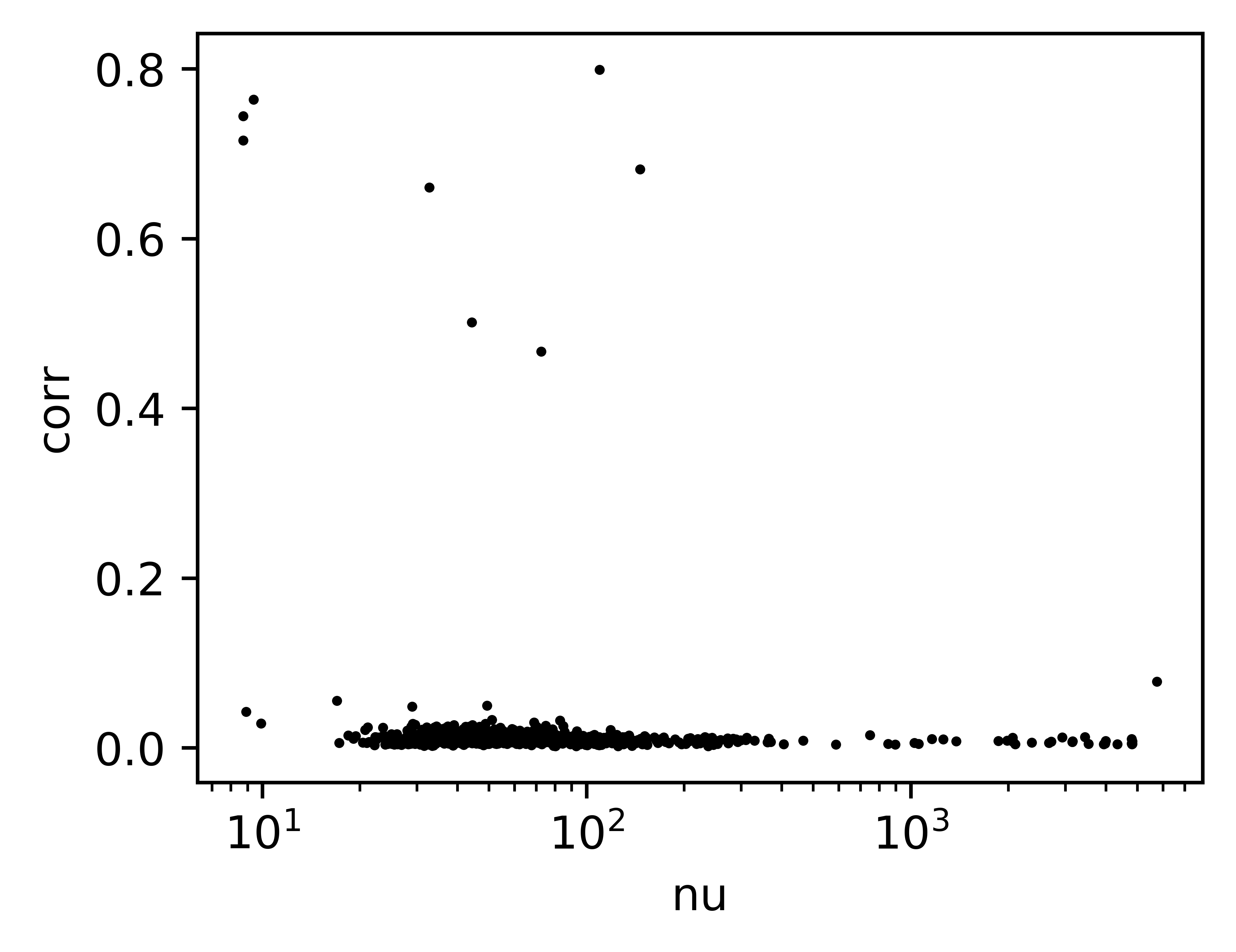}}
            \caption{Correlation $\nicefrac{\rho^d}{v^d}$ versus degrees of freedom $\nu^d$ for every $d$. Degrees of freedom on the x-axis are plotted on a logarithmic scale. \textit{Left}: Omniglot model. \textit{Middle}: CIFAR-10 model \textit{Right}: Non-convolutional version of BRUNO trained on MNIST.}
            \label{fig:dfs}
        \end{figure}

    For $\mathcal{TP}$-based models, degrees of freedom $\nu^d$ for every process in the latent space were intialized to 1000, which makes a $\mathcal{TP}$ close to a $\mathcal{GP}$. After training, most of the dimensions retain fairly high degrees of freedom, but some can have small $\nu$'s. One can notice from Figure~\ref{fig:dfs} that dimensions with high correlation tend to have smaller degrees of freedom.

\begin{figure}[h]
        \subfigure{\includegraphics[width=0.465\linewidth]{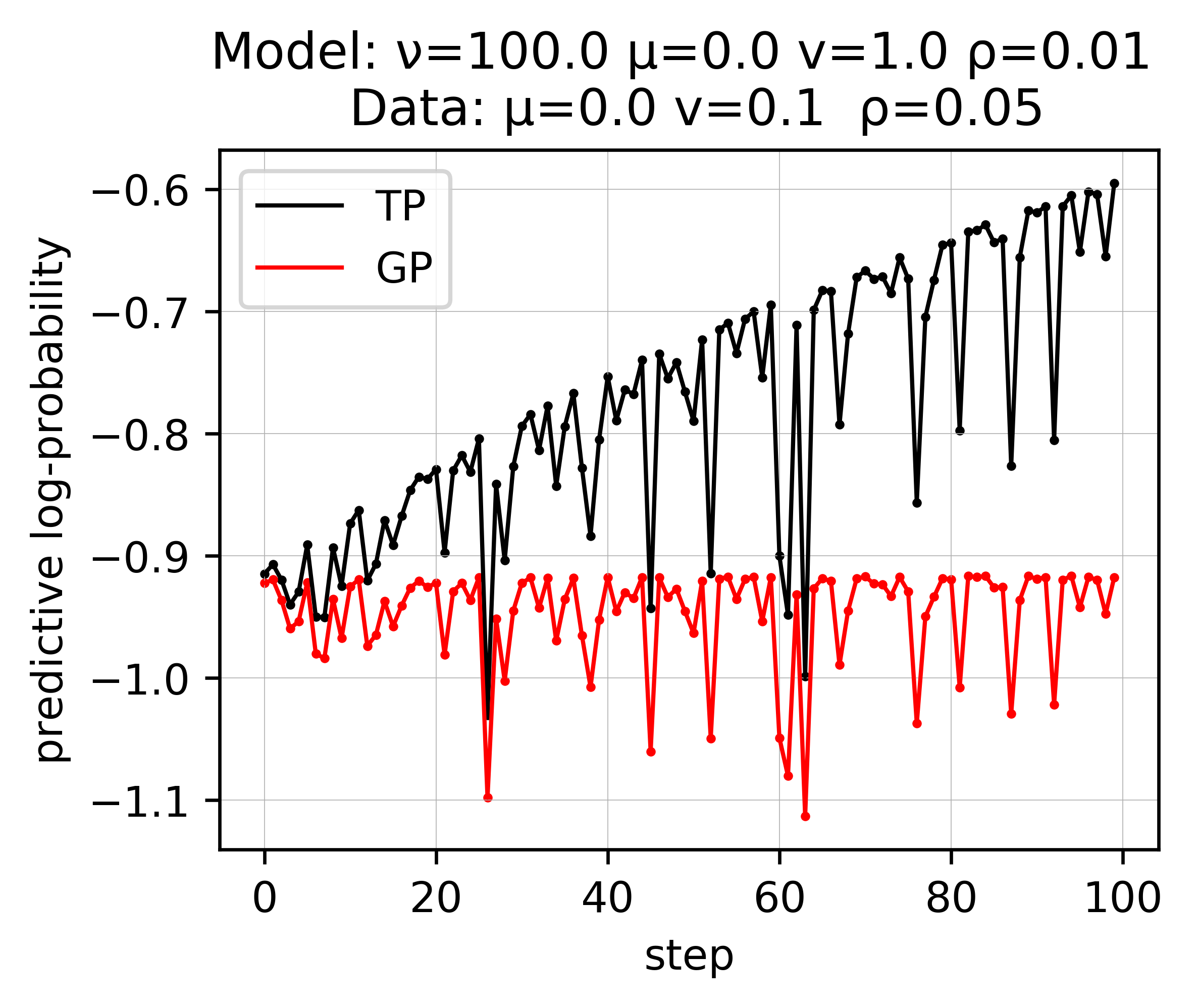}}
        \subfigure{\includegraphics[width=0.49\linewidth]{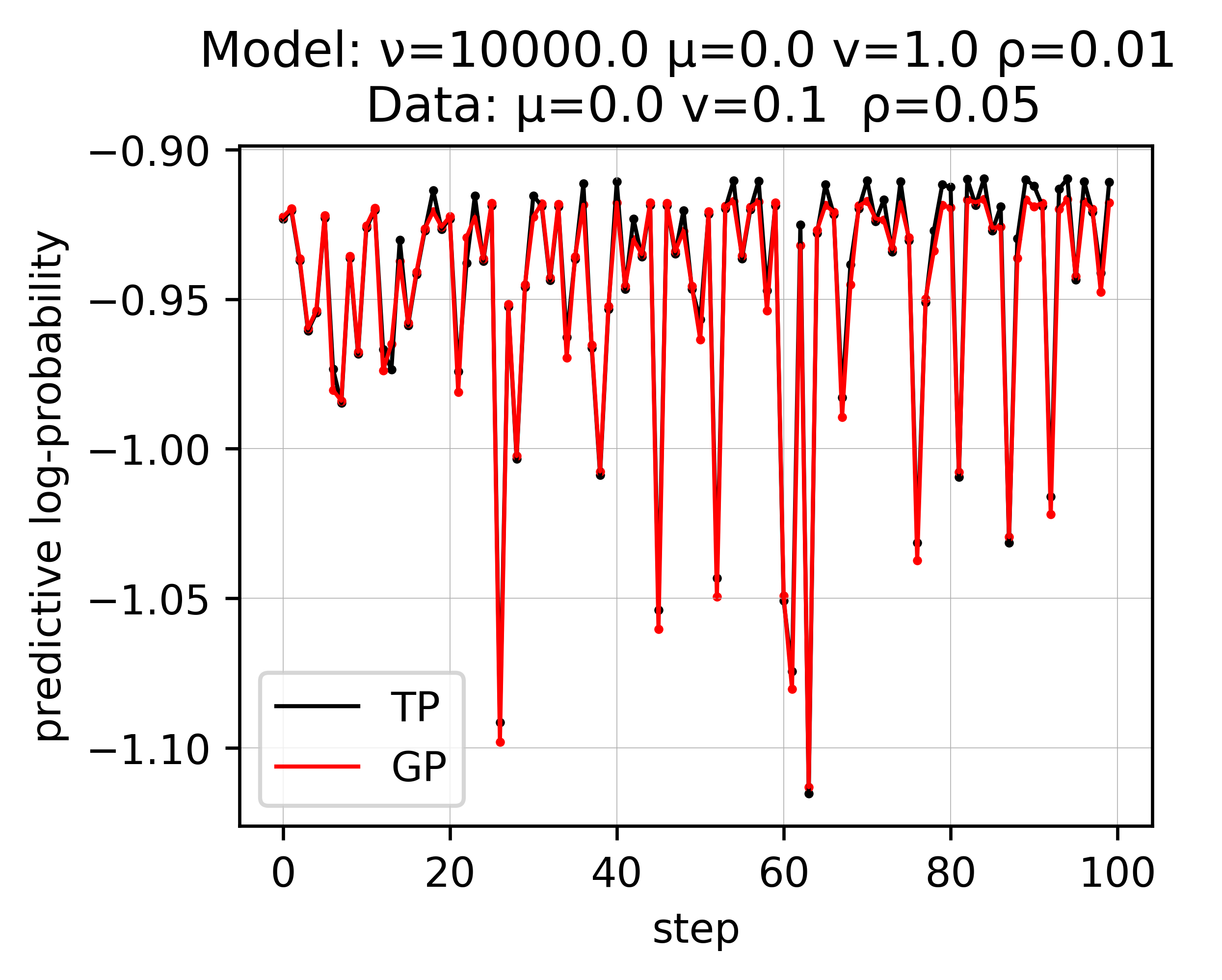}}
        \caption{ A toy example which illustrates how degrees of freedom $\nu$ affect the behaviour of a $\mathcal{TP}$ compared to a $\mathcal{GP}$. Here, we generate one sequence of 100 observations from an exchangeable multivariate normal disribution with parameters $\mu=0.$, $v =0.1$, $\rho=0.05$ and evaluate predictive probabilities under an exchangeable $\mathcal{TP}$ and $\mathcal{GP}$ models with parameters  $\mu=0.$, $v =1.$, $\rho=0.01$ and different $\nu$ for $\mathcal{TP}$s in the left and the right plots.}
        \label{fig:tp}
    \end{figure}

  We noticed that exchangeable $\mathcal{TP}$s and $\mathcal{GP}$s can behave differently for certain settings of hyperparameters even when $\mathcal{TP}$s have high degrees of freedom. Figure~\ref{fig:tp} gives one example when this is the case.

   \section{Training of $\mathcal{GP}$ and $\mathcal{TP}$-based models}

\begin{figure}[h]
        \subfigure{\includegraphics[width=0.33\linewidth]{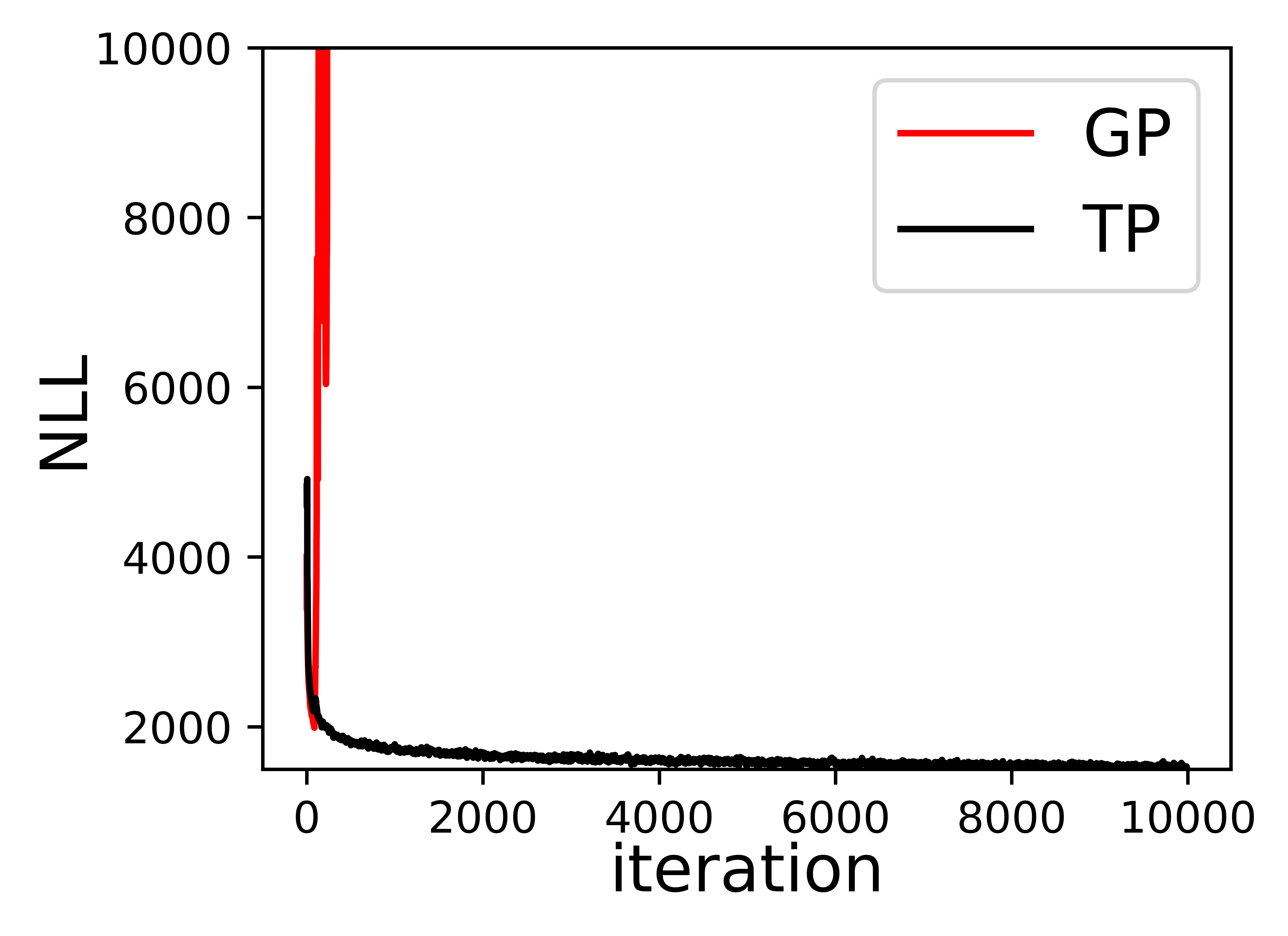}}
        \subfigure{\includegraphics[width=0.33\linewidth]{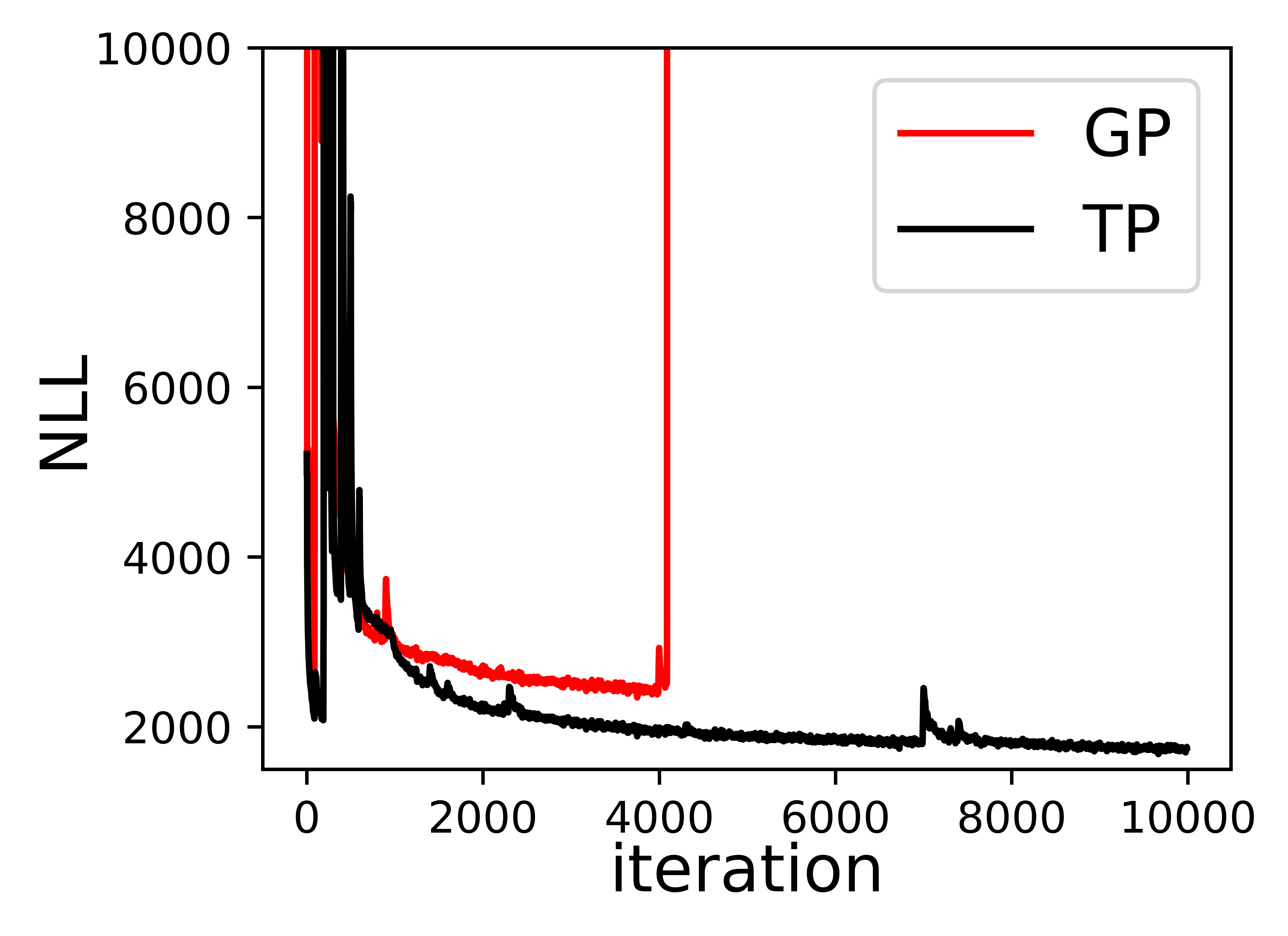}}
        \subfigure{\includegraphics[width=0.33\linewidth]{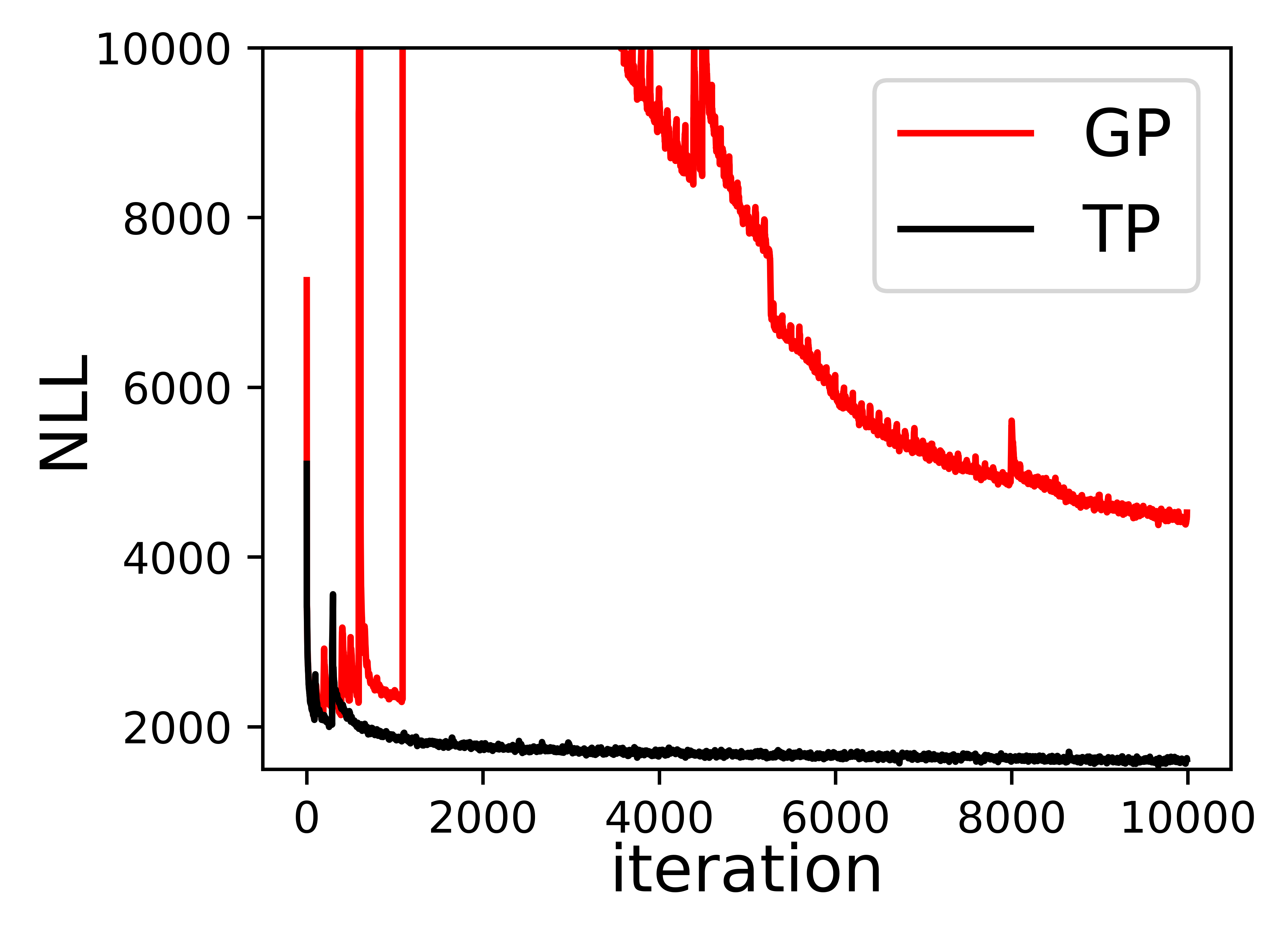}}
        \caption{Negative log-likelihood of $\mathcal{TP}$ and $\mathcal{GP}$-based BRUNO on the training batches, smoothed using a moving average over 10 points. \textit{Left:} not using weightnorm, initial covariances are sampled from $\mathcal U(0.1, 0.95)$ for every dimension. Here, the $\mathcal GP$-based model diverged after a few hundred iterations. Adding weighnorm fixes this problem. \textit{Middle:} using weightnorm, covariances are initialised to 0.1, learning rate is 0.002 (two times the default one). In this case, the learning rate is too high for both models, but the $\mathcal{GP}$-based model suffers from it more. \textit{Right:} using weightnorm, covariances are initialised to 0.95.}
        \label{fig:lr}
    \end{figure}

   When jointly optimizing Real NVP with a $\mathcal{TP}$ or a $\mathcal{GP}$ on top, we found that these two versions of BRUNO  occasionally behave differently during training. Namely, with $\mathcal{GP}$s the convergence was harder to achive. We could pinpoint a few determining factors: \textbf{(a)} the use of weightnorm~\cite{salimans16} in the Real NVP layers, \textbf{(b)} an intialisation of the covariance parameters, and  \textbf{(c)} presence of outliers in the training data. In Figure~\ref{fig:lr}, we give examples of learning curves when BRUNO with $\mathcal{GP}$s tends not to work well. Here, we use a convolutional architecture and train on Fashion MNIST. To simulate outliers, every 100 iterations we feed a training batch where the last image of every sequence in the batch is completely white.

We would like to note that there are many settings where both versions of BRUNO diverge or they both work well, and that the results of this partial ablation study are not sufficient to draws general conclusions. However, we can speculate that when extending BRUNO to new problems, it is reasonable to start from a $\mathcal{GP}$-based model with weightnorm, small initial covariances, and small learning rates. However, when finding a good set of hyperparameters is difficult, it might be worth trying the $\mathcal{TP}$-based BRUNO.

    \section{Set anomaly detection}

    Online anomaly detection for exchangeable data is one of the application where we can use BRUNO. This problem is closely related to the task of content-based image retrieval, where we need to rank an image $\bm x$ on how well it fits with the sequence $\bm x_{1:n}$~\cite{heller06}. For the ranking, we use the probabilistic score proposed in Bayesian sets~\cite{ghahramani06}:%
    \begin{equation}
        \text{score}(\bm x) = \frac{p(\bm x| \bm x_{1:n})}{p(\bm x)}.
    \end{equation}%
    When we care exclusively about comparing ratios of conditional densities of $\bm x_{n+1}$ under different sequences $\bm x_{1:n}$, we can compare densities in the latent space $\mathcal Z$ instead. This is because the Jacobian from the change of variable formula does not depend on the sequence we condition on.

\begin{figure}[htb]
        \subfigure{\includegraphics[width=0.45\linewidth]{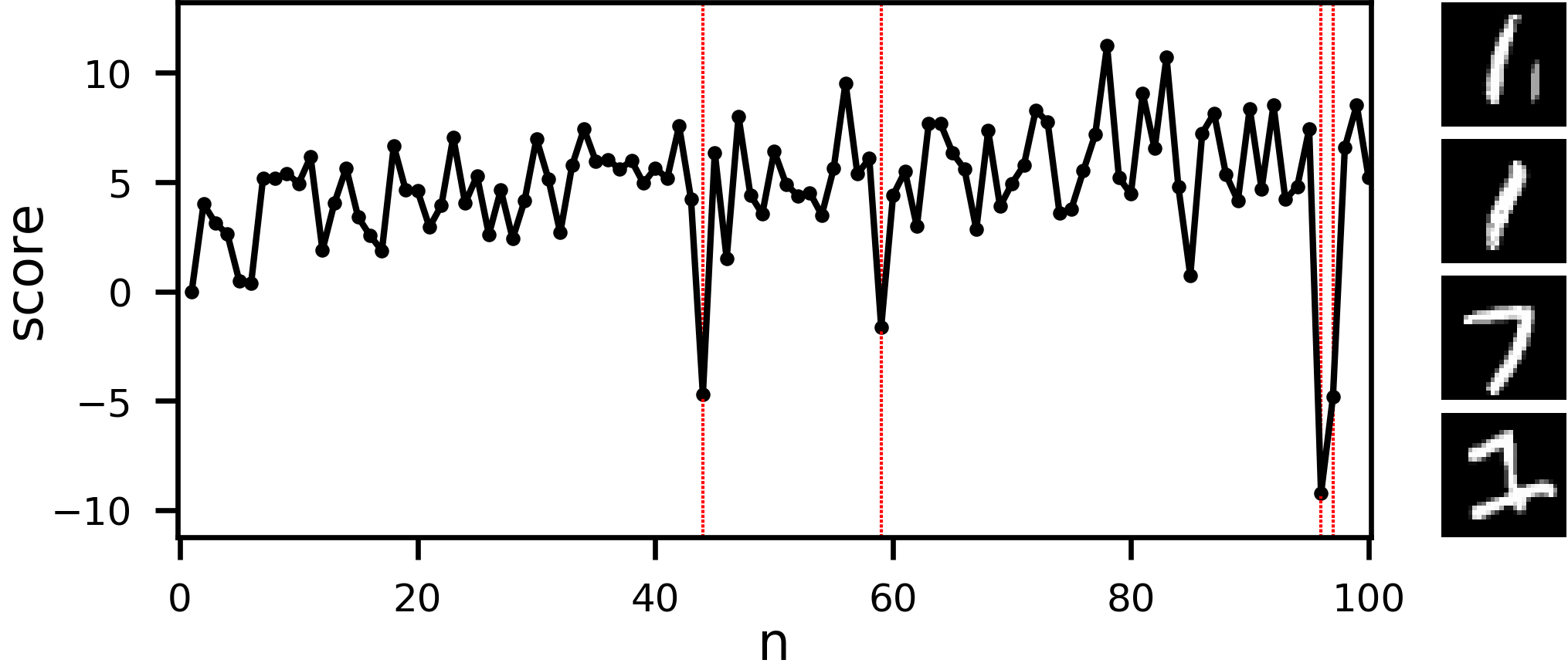}}
        \hskip 0.2in
        \subfigure{\includegraphics[width=0.45\linewidth]{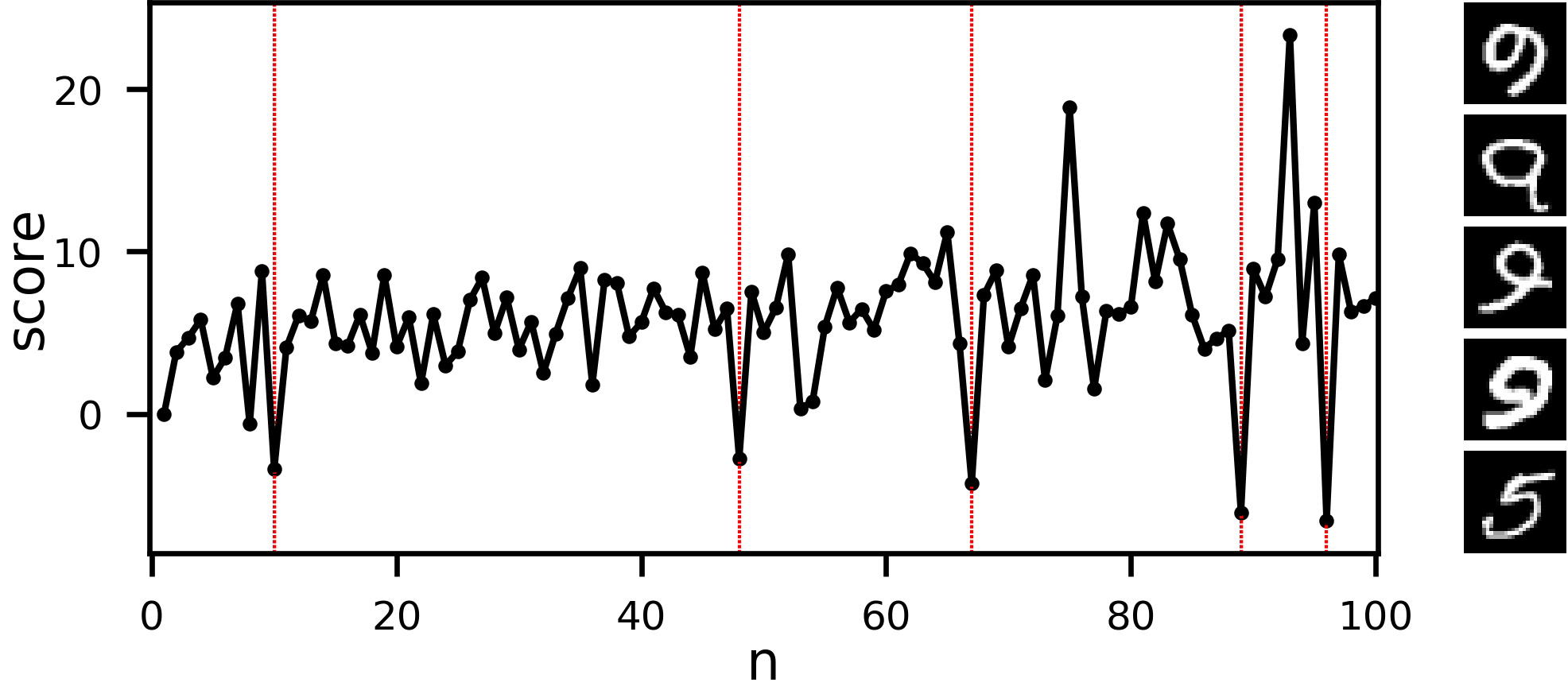}}
        \caption{Evolution of the score as the model sees more images from an input sequence. Identified outliers are marked with vertical lines and plotted on the right in  the order from top to bottom. Note that the model was trained only on images of even digits. \textit{Left:} a sequence of digit `1' images with one image of `7' correctly identified as an outlier.  \textit{Right:} a sequence of digit `9' with one image of digit `5'. }
        \label{fig:7}
    \end{figure}

    For the following experiment, we trained a small convolutional version of BRUNO only on even MNIST digits (30,508 training images). In Figure~\ref{fig:7}, we give typical examples of how the score evolves as the model gets more data points and how it behaves in the presence of inputs that do not conform with the majority of the sequence. This preliminary experiment shows that our model can detect anomalies in a stream of incoming data.

    \clearpage
    \newpage
    \section{Model samples}

    \begin{figure}[h]
        \vskip -0.1in
        \subfigure{\includegraphics[width=0.5\linewidth]{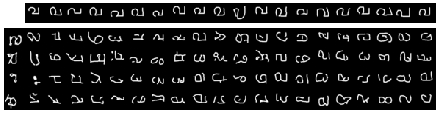}}
%        \subfigure{\includegraphics[width=0.5\linewidth]{img/sample_test_0_0.png}}
        \subfigure{\includegraphics[width=0.5\linewidth]{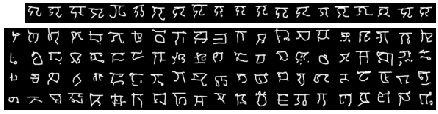}}
        \subfigure{\includegraphics[width=0.5\linewidth]{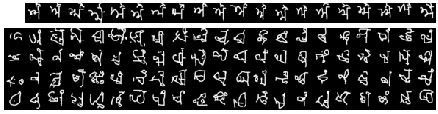}}
        \subfigure{\includegraphics[width=0.5\linewidth]{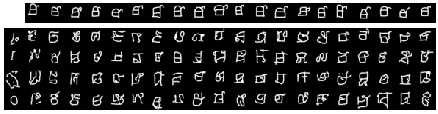}}
        \subfigure{\includegraphics[width=0.5\linewidth]{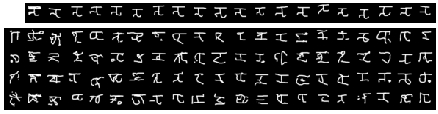}}
        \subfigure{\includegraphics[width=0.5\linewidth]{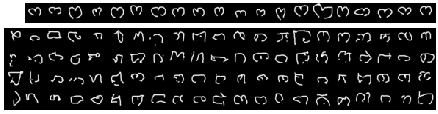}}
        \subfigure{\includegraphics[width=0.5\linewidth]{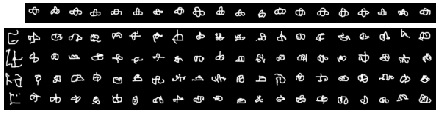}}
        \subfigure{\includegraphics[width=0.5\linewidth]{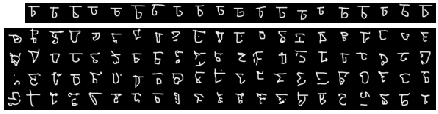}}
        \subfigure{\includegraphics[width=0.5\linewidth]{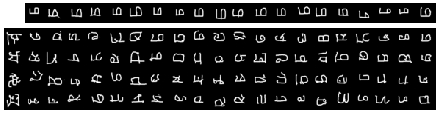}}
        \subfigure{\includegraphics[width=0.5\linewidth]{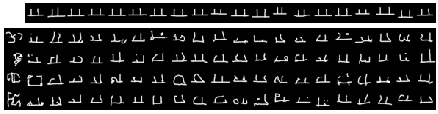}}
%        \subfigure{\includegraphics[width=0.5\linewidth]{img/sample_test_168_0.png}}
        \subfigure{\includegraphics[width=0.5\linewidth]{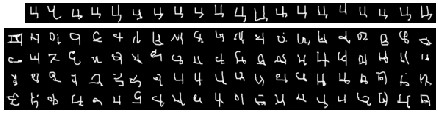}}
        \subfigure{\includegraphics[width=0.5\linewidth]{img/sample_test_188_0.png}}
        \subfigure{\includegraphics[width=0.5\linewidth]{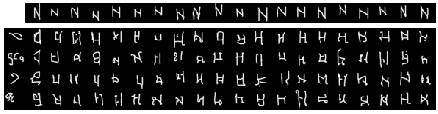}}
        \subfigure{\includegraphics[width=0.5\linewidth]{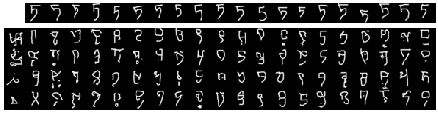}}
        \caption{Samples from a model trained on Omniglot. Conditioning images come from character classes that were not
        used during training, so when $n$ is small, the problem is equivalent to a few-shot generation.}
    \end{figure}

    \newpage
    \begin{figure}[h]
        \subfigure{\includegraphics[width=0.5\linewidth]{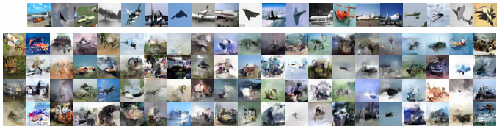}}
%        \subfigure{\includegraphics[width=0.5\linewidth]{img/cifar/sample_test_1_0.png}}
%        \subfigure{\includegraphics[width=0.5\linewidth]{img/cifar/sample_test_2_0.png}}
        \subfigure{\includegraphics[width=0.5\linewidth]{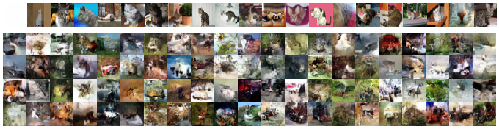}}
%        \subfigure{\includegraphics[width=0.5\linewidth]{img/cifar/sample_test_4_0.png}}
        \subfigure{\includegraphics[width=0.5\linewidth]{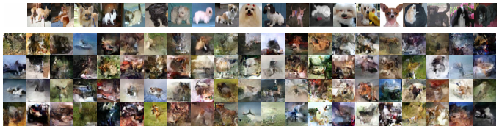}}
%        \subfigure{\includegraphics[width=0.5\linewidth]{img/cifar/sample_test_6_0.png}}
        \subfigure{\includegraphics[width=0.5\linewidth]{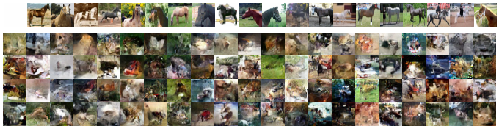}}
        \subfigure{\includegraphics[width=0.5\linewidth]{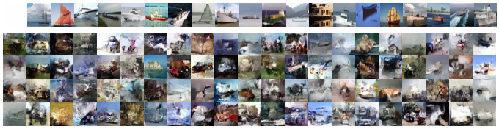}}
        \subfigure{\includegraphics[width=0.5\linewidth]{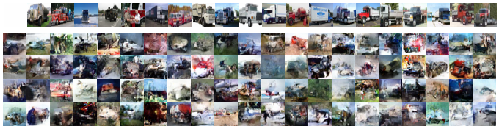}}
        \caption{Samples from a model trained on CIFAR-10. The model was trained on the set with 10 classes. Conditioning images in the top row of each subplot come from the test set.}
    \end{figure}

    \begin{figure}[h]
        \subfigure{\includegraphics[width=0.5\linewidth]{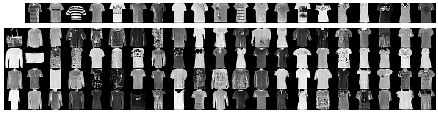}}
        \subfigure{\includegraphics[width=0.5\linewidth]{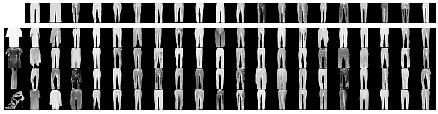}}
%        \subfigure{\includegraphics[width=0.5\linewidth]{img/fashion/sample_test_2_0.png}}
        \subfigure{\includegraphics[width=0.5\linewidth]{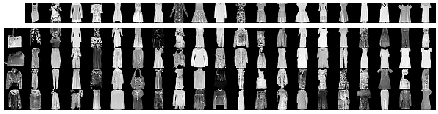}}
        \subfigure{\includegraphics[width=0.5\linewidth]{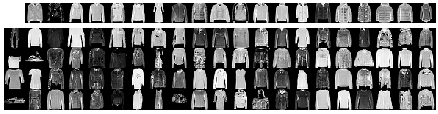}}
%        \subfigure{\includegraphics[width=0.5\linewidth]{img/fashion/sample_test_5_0.png}}
        \subfigure{\includegraphics[width=0.5\linewidth]{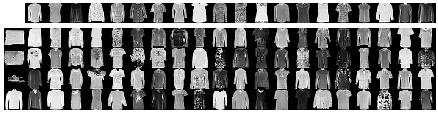}}
        \subfigure{\includegraphics[width=0.5\linewidth]{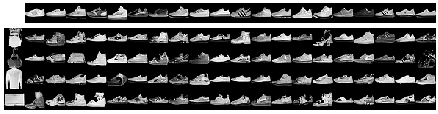}}
        \subfigure{\includegraphics[width=0.5\linewidth]{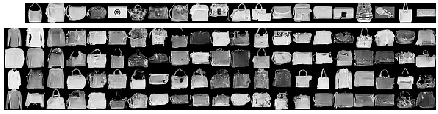}}
        \subfigure{\includegraphics[width=0.5\linewidth]{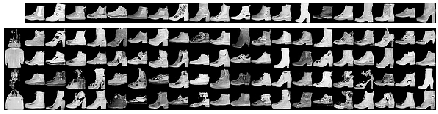}}
        \caption{Samples from a convolutional BRUNO model trained on Fashion MNIST. The model was trained
        on the set with 10 classes. Conditioning images in the top row of each subplot come from the test set.}
    \end{figure}

    \newpage
    \begin{figure}[h]
        \subfigure{\includegraphics[width=0.5\linewidth]{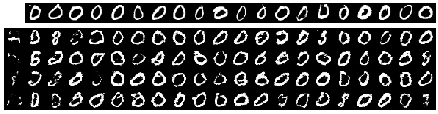}}
        \subfigure{\includegraphics[width=0.5\linewidth]{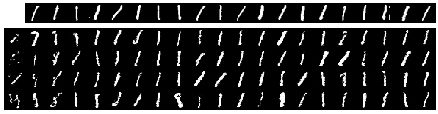}}
        \subfigure{\includegraphics[width=0.5\linewidth]{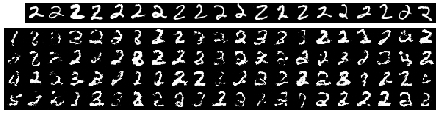}}
        \subfigure{\includegraphics[width=0.5\linewidth]{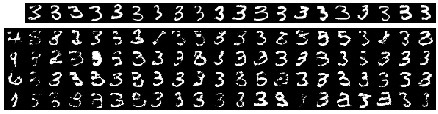}}
        \subfigure{\includegraphics[width=0.5\linewidth]{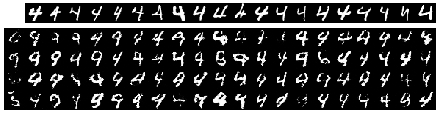}}
        \subfigure{\includegraphics[width=0.5\linewidth]{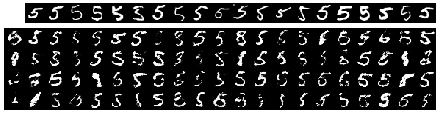}}
        \subfigure{\includegraphics[width=0.5\linewidth]{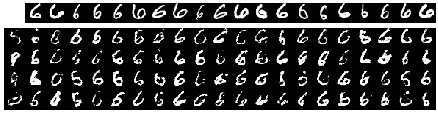}}
        \subfigure{\includegraphics[width=0.5\linewidth]{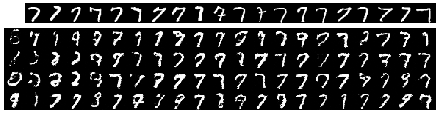}}
        \subfigure{\includegraphics[width=0.5\linewidth]{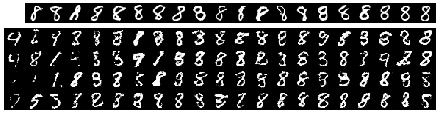}}
        \subfigure{\includegraphics[width=0.5\linewidth]{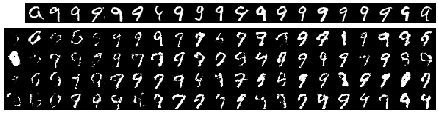}}
        \caption{Samples from a non-convolutional model trained on MNIST. The model was trained
        on the set with 10 classes. Conditioning images in the top row of each subplot come from the test set.}
    \end{figure}

\end{document}